%% file: main.tex
\documentclass[journal]{IEEEtran}

\usepackage{todonotes}

\usepackage{moreverb,url}

\usepackage{amsmath,amsfonts}
\usepackage{array}
\usepackage[caption=false,font=normalsize,labelfont=sf,textfont=sf]{subfig}
\usepackage{textcomp}
\usepackage{stfloats}
\usepackage{url}
\usepackage{verbatim}
\usepackage{graphicx}
\usepackage{cite}

\def\BibTeX{{\rm B\kern-.05em{\sc i\kern-.025em b}\kern-.08em
    T\kern-.1667em\lower.7ex\hbox{E}\kern-.125emX}}
\usepackage{balance}

\usepackage{siunitx}

\usepackage{algpseudocode}

\usepackage{algorithm}
\usepackage{caption}
\algrenewcommand{\algorithmiccomment}[1]{\textcolor{gray}{\hskip1em// #1}}
\algnewcommand{\LineComment}[1]{\State\textcolor{gray}{// #1}}
\algrenewcommand\algorithmicrequire{\textbf{Input:}}
\algrenewcommand\algorithmicensure{\textbf{Output:}}

\DeclareMathOperator*{\argmin}{arg\,min}

\usepackage{hyperref}

\newcommand{\etal}{\textit{et~al.}}
 
\begin{document}

\title{Multiple-Hypothesis Path Planning with Uncertain Object Detections}

\author{Brian H. Wang, Beatriz Asfora, Rachel Zheng, Aaron Peng, Jacopo Banfi, and Mark Campbell

\thanks{\emph{Corresponding author: Brian H. Wang}}
\thanks{Brian H. Wang, Beatriz Asfora, Rachel Zheng, Aaron Peng, and Mark Campbell are with the Sibley School of Mechanical and Aerospace Engineering, Cornell University, Ithaca, NY 14850, USA (e-mail: bhw45@cornell.edu; ba386@cornell.edu; rz246@cornell.edu; ahp67@cornell.edu; mc288@cornell.edu).}
\thanks{Jacopo Banfi is with the  Computer Science and Artificial Intelligence Laboratory, Massachusetts Institute of Technology, Cambridge, MA 02139, USA (e-mail: jbanfi@mit.edu).}
}

\markboth{}%
{Wang \etal{}: Multiple-hypothesis path planning with uncertain object detections}

\maketitle

\input{abstract}


\input{definitions}

\input{introduction}
\input{related_work}
\input{approach}
\input{experiments}
\input{conclusion}

\bibliographystyle{IEEEtran}
\bibliography{references}

\end{document}

%% file: abstract.tex
\begin{abstract}
%
Path planning in obstacle-dense environments is a key challenge in robotics, and depends on inferring scene attributes and associated uncertainties. We present a multiple-hypothesis path planner designed to navigate complex environments using obstacle detections. Path hypotheses are generated by reasoning about uncertainty and range, as initial detections are typically at far ranges with high uncertainty, before subsequent detections reduce this uncertainty. Given estimated obstacles, we build a graph of pairwise connections between objects based on the probability that the robot can safely pass between the pair. The graph is updated in real time and pruned of unsafe paths, providing probabilistic safety guarantees. The planner generates path hypotheses over this graph, then trades between safety and path length to intelligently optimize the best route. We evaluate our planner on randomly generated simulated forests, and find that in the most challenging environments, it increases the navigation success rate over an A* baseline from 20\% to 75\%. Results indicate that the use of evolving, range-based uncertainty and multiple hypotheses are critical for navigating dense environments. 
\end{abstract}

%% file: definitions.tex


\newcommand{\est}{\hat}
\newcommand{\rvar}{}
\newcommand{\vc}{\vec}

\newcommand{\tree}{i}
\newcommand{\bb}{j}
\newcommand{\tme}{k}


\def\obsgt{\vec{O}}  
\def\allobsgt{\boldsymbol{O}}  
\def\ntreesgt{n_{\rm obs}}
\def\treed{d_{\tree}}

\def\treex{x_{\tree}}
\def\treey{y_{\tree}}
\def\treed{d_{\tree}}

\def\detected{\boldsymbol{D}}

\def\treemeasvec{\vec{z}_{\tree, \tme}^{~\rm bb}}
\def\treexmeas{{x}_{\tree, \tme}^{\rm bb}}
\def\treeymeas{{y}_{\tree, \tme}^{\rm bb}}
\def\treedmeas{{d}_{\tree, \tme}^{\rm bb}}

\def\treerb{\vec{z}_{\tree, \tme}} 
\def\treerbr{{r}_{\tree, \tme}}
\def\treerba{{\phi}_{\tree, \tme}}
\def\treerbd{{d}_{\tree, \tme}}

\def\noiserange{W_{\tree, \tme}^{r}}
\def\noisebearing{W_{\tree, \tme}^{\phi}}
\def\noisediam{W_{\tree, \tme}^{d}}

\def\varnoiserange{W_{\tree, \tme}^{r}}
\def\varnoisebearing{W_{\tree, \tme}^{\phi}}
\def\varnoisediam{W_{\tree, \tme}^{d}}

\def\btreerbr{\bar{{r}}_{\tree, \tme}}
\def\btreerba{\bar{{\phi}}_{\tree, \tme}}
\def\btreerbd{\bar{{d}}_{\tree, \tme}}

\def\rot{R_{ij}}

\def\obsmu{\vec{\mu}_{O_i}}
\def\obssigma{\Sigma_{O_i}}
\def\sigmax{\sigma_{xx, i}}
\def\sigmay{\sigma_{yy, i}}
\def\sigmad{\sigma_{dd, i}}

\def\rangemax{r_{\rm max}}
\def\delaunay{\rm DT}
\newcommand{\vtx}[1]{v_{\rm #1}}
\def\localwaypoints{W_{\rm L}}
\def\state{\vec{q}}

\def\N{\mathcal{N}}

\def\start{\state_{start}}  
\def\globalgoal{\vec{x}_{\rm G}}
\def\globalgoalxy{\left(x_{\rm G}, y_{\rm G}\right)}
\def\localgoal{\vec{x}_{\rm L}}    
\def\localgoalxy{\left(x_{\rm L}, y_{\rm L}\right)}


\def\lm{\ell} 
\def\lmest{\est{\lm}} 

\def\nlm{n_{\rm LM}}  

\def\obs{\hat{O}}  
\def\obsxy{\hat{\vec{x}}}  
\def\obsd{\hat{d}}  
\def\allobs{\hat{\boldsymbol{O}}}  
\def\nobs{n_{\rm obs}}  
\newcommand{\obsstate}[1]{\left\{x_#1, y_#1, d_#1\right\}} 
\def\sigsq{\sigma^2}  
\def\wr{w_R}  

\def\meanvec{\vec{\mu}}

\newcommand{\navprob}[2]{P_{\rm N}\left(#1, #2\right)}

\def\nonneg{\mathbb{R}_{\geq0}}
\def\edt{E_{\rm DT}}  
\def\vdt{V_{\rm DT}}  
\def\tdt{T_{\rm DT}}  
\def\cellface{e_{\rm DT,ij}}  


\def\graphname{navigation graph} 
\def\gnav{G_{\rm N}}  
\def\enav{E_{\rm N}}  
\def\vnav{V_{\rm N}}  
\def\hyp{H}       
\def\setofranges{\mathcal{R}}     
\def\range{R}     
\def\path{P}       
\def\costdist{c_{\rm dist}} 
\def\costsafe{c_{\rm safe}} 
\def\probsafe{p_{\rm safe}} 

\def\rangeshort{r_{\rm short}}  
\def\rangemed{R_{2}}  
\def\safetyshort{p_{\rm min}}   
\def\desiredprob{p_{\rm target}} 
\def\minprob{p_{\rm min}} 
\def\queue{\mathcal{Q}}    
\def\paths{\mathcal{P}}    

\def\nhyp{N_{\rm hyp}}  
\def\wdist{\alpha_{\rm dist}}  
\def\wsafe{\alpha_{\rm safe}}  

\def\priority{\pi}

\def\pathcostdist{C_{\rm dist}}
\def\pathcostsafe{C_{\rm safe}}
\def\pathcosttotal{C_{\rm total}}

\def\bestpath{\path^*}
\def\planaheaddist{d_{\rm local}}

\def\detrange{R_{\rm det}}  
\def\mapwidth{W_{\rm M}}
\def\mapheight{H_{\rm M}}
\def\mapshape{(\mapwidth \times \mapheight)}

\def\forestdensity{\rho}

%% file: introduction.tex

\section{Introduction}

\begin{figure}
    \centering
    \includegraphics[width=\columnwidth]{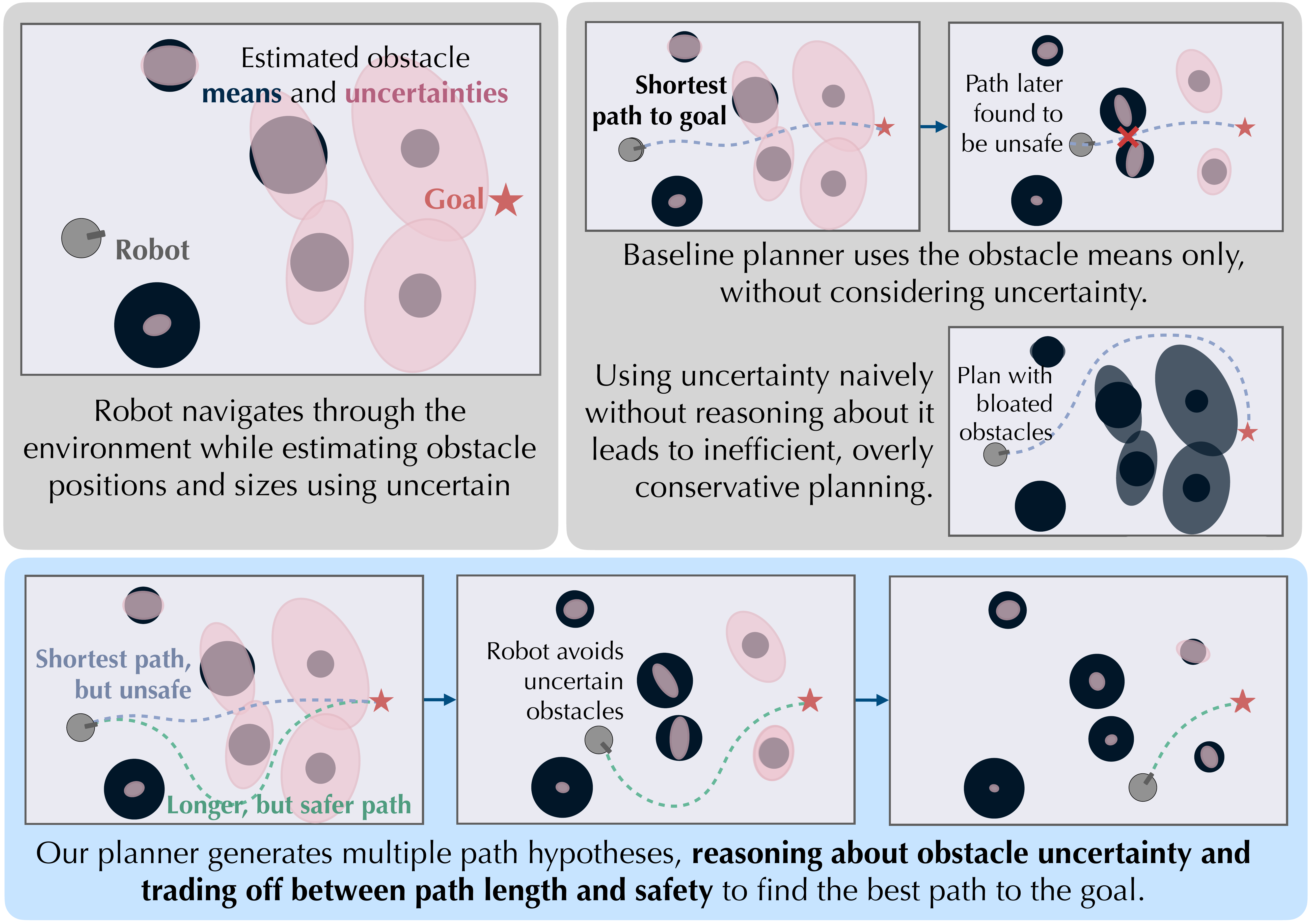}
    \caption{Overview of the contributions of our multiple-hypothesis planner.}
    \label{fig:mhp_overview}
\end{figure}

Autonomous navigation in unknown, unstructured, and obstacle-dense environments requires a robot to recognize obstacles and plan a safe path around them, typically by using a 3D sensor such as a stereo camera or lidar. This problem becomes much more challenging as the desired navigation speed and complexity of the environment increase. The further ahead the robot can robustly perceive and react to upcoming obstacles, the more safely it will be able to navigate through increasingly challenging and obstacle-dense environments.

However, 3D sensors used in robotics are often affected by measurement errors that increase with distance from the sensor. Stereo camera depth errors grow quadratically with range, due to disparity estimation errors \cite{wang2019pseudo, monica2020vision}. Lidar sensors, while more precise, return very sparse points at longer ranges. 

Current pipelines for autonomous navigation in unknown environments plan iteratively,  by constructing a map of the robot's immediate surroundings (often as a 3D occupancy grid), optimizing a path to a global goal in this map, then recursively updating the local map and planned trajectory as the robot moves towards the goal and obtains new sensor measurements \cite{ryll2019efficient, mohta2018fast}. These methods have demonstrated significant successes in autonomous path planning. However, due to increasing 3D sensor noise with range, existing methods typically plan only in a short range near the robot, where sensor noise is minimal and reliable map construction is possible. The robot discards noisier sensor measurements beyond this range entirely, ultimately limiting the robot's ability to quickly and effectively traverse complex environments.

Ryll \etal{} \cite{ryll2020semantic} point out this constrained perception horizon as a key limitation on the speed of safe autonomous navigation in agile unmanned aerial vehicle (UAV) flight. With no knowledge of the environment past the short, minimal-noise sensor range, the robot must plan conservatively, always ready to come to a stop at the boundary of this sensor range in case new obstacles appear. Ideally, the robot should be able to reason about potential obstacles at far ranges, enabling preemptive avoidance of obstacle-dense areas.

Long-range sensor measurements, while noisy, still contain usable information for recognizing obstacles. Works such as \cite{ryll2020semantic}, \cite{yutaoDeepSemanticHPPC}, and \cite{shaban2021semantic} presented planning methods that use camera image semantics to augment short range 3D sensor measurements. Recent works in 3D object detection \cite{wang2019pseudo, you2019pseudo, shi2019pointrcnn, lang2019pointpillars, qian2020end} have demonstrated accurate object detection at longer ranges than reliable occupancy grid mapping is typically possible. These methods use machine learning to recognize patterns in 3D point clouds and detect objects even within noisy sensor data.

These object detectors provide a promising way to extend the amount of actionable information available to a robot, by enabling longer-range obstacle recognition. However, object detections still contain errors in measuring obstacle positions and sizes, as shown previously in \cite{wang2021detectingtrees}. The robot must therefore reason about this sensing uncertainty in the path planning algorithm, in order to effectively use these object detections. 

As illustrated in Figure \ref{fig:mhp_overview}, failing to consider this uncertainty can result in unsafe planned paths, while using uncertainty naively can result in inefficient planning. In this paper, we present a multiple-hypothesis planner which uses uncertain object detections to help the robot navigate to its goal, while considering the length and safety of multiple candidate paths to determine the overall best path to the goal. 

Our work is motivated by navigation through a forest while avoiding tree trunks, previously studied in  \cite{karaman2012limitedsensingrrange}, \cite{choudhury2015theoretical}, and \cite{junior2021lattice}. To model this problem, we consider a robot navigating through an environment densely populated with obstacles of a single class, but with different sizes and locations, that can be detected with noise that varies with range. We first estimate the positions and sizes of obstacles using these noisy object detections. We then construct a graph representation of the scene based on pairwise connections between obstacles, which stores the probability that the robot can safely navigate between each pair of nearby obstacles in the world. This graph can be updated in real-time based on new measurements, and pruned if edges do not provide safe paths, thus providing probabilistic guarantees. Using this probabilistic graph, we plan multiple path hypotheses between pairs of obstacles to the navigation goal. The planner then intelligently optimizes the best route to the goal by trading off between safety and expected path length.

Our contributions include the following:

\begin{itemize}
    \item A high-level graph representation of forest environments, that represents and stores the probabilities of the robot being able to safely navigate between pairs of obstacles in the forest.
    \item A multiple-hypothesis path planning method which uses this probabilistic navigation graph to generate and evaluate multiple candidate paths to the goal, accounting for both expected path distance and safety.
    \item An experimental simulation study in randomly generated forest environments, demonstrating that our probabilistic model of long-range object detection, when compared to a baseline planner using a 2D A* search, allows a robot to more safely reach its goal. In the most challenging and obstacle-dense forests, while the A* baseline is able to reach the goal in only 20\% of environments, our planner increases the navigation success rate to 75\%.
    \item An open-source, publicly available implementation of our planner and simulation, available at \textbf{\url{https://github.com/brian-h-wang/multiple-hypothesis-planner}}.
\end{itemize}

%% file: related_work.tex

\section{Related work}

\subsection{Path planning in unknown environments}

Traditionally, path planning methods take as input a known map of the robot's surroundings, then use this map to plan a collision-free path to a given goal state. Robot path planning methods include sampling-based approaches such as rapidly-exploring random trees (RRTs) \cite{lavalle2001rrt, kuwata2009real}, as well as minimum cost search methods, often based on the A* algorithm \cite{hart1968formal}. Hybrid A* \cite{dolgov2010pathplanning,petereit2012hybridastar} extends this algorithm by associating each discrete step of the A* search with a continuous vehicle state, ensuring that the final path is valid with respect to the robot kinematics.

In practice, a robot generally lacks a detailed map of the environment in advance and must construct one in real time using its onboard sensors. Therefore, typical path planning pipelines work iteratively, re-planning as the robot traverses the environment and updates its map with new sensor measurements. A commonly used map structure is a binary 3D occupancy grid, where cells are classified as either occupied or unoccupied. These occupancy grids can be constructed using 3D sensors such as stereo cameras \cite{ryll2019efficient} or lidar \cite{mohta2018experiments, mohta2018fast, tian2018search}. These works have significantly advanced the state of the art in 3D autonomous path planning, implementing the iterative re-planning approach on 3D grids.

A key limiting factor on autonomous navigation is the maximum perception horizon of the robot's sensors \cite{ryll2020semantic}. At long ranges, sensor noise grows significantly, with stereo cameras suffering from a quadratic increase in depth estimation error with range \cite{wang2019pseudo,monica2020vision}, and lidar point measurements becoming sparse. Longer range planning carries computational costs as well; for example, the cost to maintain a 3D occupancy grid grows cubically with the size of the grid, forcing size- and weight-constrained robots to use a smaller grid that does not make use of all available sensor information \cite{mohta2018fast}. 

Planning only at short range limits the speed of safe robotic navigation, as the robot must move slowly enough to be able to come to a stop at the boundary of the perceived space, in case new obstacles are detected. \cite{tordesillas2021faster} address this issue by planning in the unknown space, allowing faster flight while maintaining a safe backup trajectory that exists entirely in the nearer-range known space. However, the planner in this work assumes the unknown space to always be free. Ideally, a robot should be able to reason about uncertainty in noisier, longer-range sensor measurements, and plan efficient paths that also account for the imperfect knowledge of distant regions. 

Additionally, complex and cluttered environments make efficient and agile planning especially difficult. Banfi \etal{} \cite{banfi2022occupancygrid} previously suggested that reasoning about multiple path hypothesis could be beneficial when planning in uncertain environments. However, the map model used in that work is based on a 3D occupancy grid, whose typical resolutions (15-30 cm, to enable real-time processing) might preclude the existence of safe paths in dense environments. The present work fills this gap.

Other previous works have specifically addressed navigation in forests, a common outdoors environment which demonstrates the challenges of unstructured and obstacle-dense environments. \cite{karaman2012forestflight} and \cite{choudhury2015theoretical} develop theoretical guarantees on the speed of flight through a forest modelled as a Poisson point process, with the former additionally analyzing the effects of limiting the robot's sensing range. \cite{li2020localization} and \cite{nardari2020place} develop methods for 2D localization in forests, addressing the challenges of obstacle-dense environments with limited semantic information differentiating tree trunk obstacles. \cite{junior2021lattice} present a computationally efficient method for path planning in a forest, by constructing a lattice of candidate paths, then pruning it according to detected obstacle locations. \cite{tian2018search} present a forest search and rescue system, mapping a UAV's surroundings using a hybrid approach including a 3D occupancy grid along with a data-efficient object-based representation of tree trunks, then planning paths using A* search. This paper builds upon previous works by reasoning about the \emph{uncertainty} in tree trunk obstacle estimates, when planning paths through the forest. 

\subsection{Perception for planning}

Robotics researchers have begun using machine learning-based perception to enable robots to more meaningfully understand their surroundings. \cite{bowman2017probabilistic}, \cite{nicholson2018quadricslam}, and \cite{ok2019robust} use 2D bounding box object detections as measurements for landmark-based simultaneous localization and mapping (SLAM). In contrast to previous SLAM methods, which heuristically pick out landmarks from 2D image pixels or 3D point cloud scans, the object detector enables the robot to reason at a higher level, recognizing objects of interest, estimating their positions and sizes, and localizing with respect to them. 

Researchers have also used learned perception to augment  robotic path planning and navigation. \cite{yutaoDeepSemanticHPPC} use a Bayesian neural network to identify potentially unsafe regions in stereo images (for example, distinguishing muddy terrain from a paved sidewalk), then use this information to plan safe paths that also decrease the robot's uncertainty about its surroundings, using a next-best view planning approach.  \cite{shaban2021semantic} train a neural network to predict drivable versus unsafe terrain from sparse lidar scans. \cite{lim2021class} extend A* search with semantic weights applied to the graph edges.  Finally, \cite{ryll2020semantic} use semantic labels on RGBD camera images to identify from a distance regions where a UAV can safely fly in urban environments. Ryll \etal{} \cite{ryll2020semantic} motivate their work with the specific goal of raising the maximum speed for safe UAV flight, by increasing the robot's perception range. The common theme throughout these works is that semantic information allows a robot to reason about further away regions in the environment, where traditional 3D sensors give sparser and/or noisier measurements. This capability increases the robot's effective perception range, allowing it to recognize potential unsafe areas at a further distance.

Similarly motivated, in this paper we explore the usage of 3D object detectors for path planning. Recent works in 3D object detection have shown impressive results for object detection using sensors such as lidar and stereo cameras. Lidar provides highly precise 3D point clouds which have been used successfully as inputs to 3D object detection methods \cite{shi2019pointrcnn, lang2019pointpillars}. Recently, \cite{wang2019pseudo} and \cite{you2019pseudo}, motivated by the lower cost, weight, and power consumption of stereo cameras compared to lidar, proposed a pseudo-lidar data representation that significantly closed the accuracy gap between lidar- and stereo camera-based object detection methods. 

One challenge with stereo camera 3D object detection is the fact that due to significant noise, long-range stereo point clouds cannot be annotated with 3D bounding boxes in a reliable and unbiased way. Previously, in \cite{wang2021detectingtrees}, we addressed this problem by introducing a self-supervised data labeling and detector training pipeline. Our approach fuses a series of noisy point clouds into a cleaned global point cloud, clusters obstacles in the fused point cloud, then propagates obstacle locations back onto individual stereo point clouds, using the pose history of the camera.

%% file: approach.tex


\input{approach_subsections/problem-definition}

\section{Multiple-hypothesis planning under uncertainty}

\input{approach_subsections/safety-probabilities}

\input{approach_subsections/graph-construction}

\input{approach_subsections/graph-planning}

\input{approach_subsections/guarantees}


%% file: approach_subsections/problem-definition.tex

\section{Problem definition}

\subsection{Forest environment}

As an example unstructured and obstacle-dense environment for robotic path planning, we consider the planar forest model previously studied in \cite{karaman2012forestflight}, \cite{choudhury2015theoretical}, and \cite{junior2021lattice}. The robot's objective is to navigate through the forest to a 2D goal point $\globalgoal = \globalgoalxy$ specified in a global coordinate frame, while avoiding densely distributed 2D obstacles representing tree trunks. We model tree trunks as circular obstacles, as seen from a top-down 2D perspective. The set 
\begin{equation}
 \allobsgt = \{\obsgt_i = (x_i, y_i, d_i), \forall i \in [1, \ntreesgt]\}   
 \label{eq:obs_gt}
\end{equation}
represents the actual obstacles in the environment, where $x_i$ and $y_i$ are the 2D coordinates of tree $i$ in the global reference frame, and $d_i$ is its diameter.

\subsection{Obstacle position and size estimation}\label{sec:obstacle_estimation}

The robot does not know the number of obstacles $\ntreesgt$, their locations, or their sizes ahead of time, and estimates these variables using noisy sensor measurements as it navigates.

We assume the robot estimates the 2D positions and diameters of obstacles as Gaussian-distributed random variables. We define an obstacle estimate as 
\begin{equation}\label{eq:obs_estimate}
    \obs_i \sim \mathcal{N}\left(\obsmu, \obssigma \right),
\end{equation}
where the estimate's mean state vector contains the expected 2D position and diameter of tree $i$,
\begin{equation}
    \obsmu = \begin{bmatrix}\mu_{x,i} & \mu_{y,i} & \mu_{d,i} \end{bmatrix},
\end{equation}
and $\obssigma$ is a 3D covariance matrix encoding the uncertainty in the estimate. The elements on the diagonal of $\obssigma$, which we denote $\sigmax^2$, $\sigmay^2$, and $\sigmad^2$, are the variances of the estimates of $x_i$, $y_i$, and $d_i$ respectively. Gaussian estimates in this form can be produced by various estimation algorithms such as the Kalman filter and its variants, or factor graph SLAM \cite{kaess2008isam}. 

These obstacle estimates are updated iteratively using newly received obstacle measurements. We assume that at each time step $k$, the robot detects some subset of the trees in the forest--for example, all visible trees within a certain maximum range and angular field of view, except for any of those that register as false negatives due to detector errors. We denote the set of trees detected at time step $k$ as $\detected_k \subseteq \allobsgt$.

For each tree $\tree$ in $\detected_k$, the robot receives a range, bearing and diameter measurement:
\begin{equation}
    \treerb = \begin{bmatrix} \treerbr & \treerba & \treerbd \end{bmatrix}^{T}.
\end{equation}

We model each as a Gaussian variable,
\begin{equation}
    \treerbr = \btreerbr + \noiserange\left(\btreerbr\right),
\end{equation}
\begin{equation}
    \treerba = \btreerba + \noisebearing,
\end{equation}
\begin{equation}
    \treerbd = \btreerbd + \noisediam\left(\btreerbr, \btreerbd\right),
\end{equation}
where $\btreerbr,~\btreerba$ and $\btreerbd$ are the measurement means, i.e. the true range and bearing from the robot to the obstacle, and the obstacle's true diameter. $\noiserange, \noisebearing$ and  $\noisediam$ are zero mean, Gaussian distributed random variables modeling the sensor noise on these quantities. The range measurement noise depends on the range from the robot to the obstacle, and we assume generally that the measurement variance will increase with range, as studied in \cite{monica2020vision}. We also assume that the diameter measurement noise may vary with range and with the true size of the detected obstacle.

Given these range-bearing measurements, we calculate the 2D obstacle estimates in \eqref{eq:obs_estimate} by solving a factor graph landmark SLAM problem, defining range and bearing factors between the detected trees at time step $k$ and the corresponding robot pose factor. Additionally, we estimate the obstacle sizes from the diameter measurements. 

\subsection{Local planner definition}\label{sec:local_planner}

Existing planning pipelines \cite{mohta2018fast, ryll2019efficient, tordesillas2021faster} generally include a \emph{local planner} which generates a dynamically feasible short-term trajectory for the robot. This local planner complements the global planner, which plans a high-level path that reaches all the way to the global goal, but does not necessarily account for the robot dynamics due to computational constraints.

We assume the local planner takes as input a local goal position $\localgoal = \localgoalxy$, the robot dynamics function, and a map representation, and outputs a series of 2D waypoints
\begin{equation}
    \localwaypoints = \begin{bmatrix}
    x_0 & y_0 \\
    x_1 & y_1 \\
    \vdots & \vdots \\
    x_{n_{w}} & y_{n_{w}}\\
    x_L & y_L
    \end{bmatrix}
\end{equation}
which form a path from the robot's current state $\state_0 = \begin{bmatrix}x_0 & y_0 & \theta_0 \end{bmatrix}^T$ to the goal position $\localgoal$ through $n_w$ waypoints. In the following sections of this paper, we describe a modular high-level planner which augments the local planner by providing a local goal computed using long-range, uncertain object detections.

%% file: approach_subsections/safety-probabilities.tex

\begin{figure*}
    \centering
    \includegraphics[width=\textwidth]{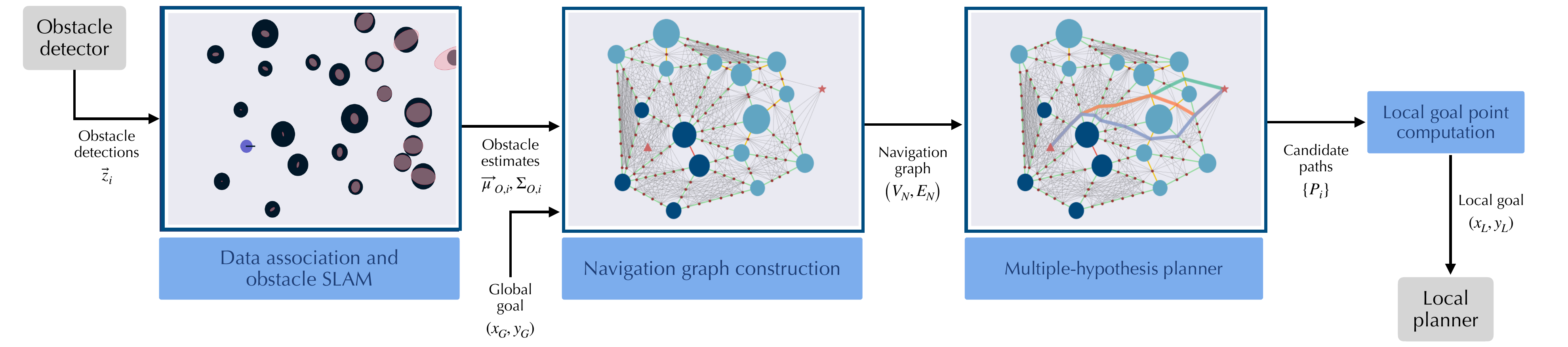}
    \caption{Illustration of the components of our high-level planning method, and their inputs and outputs.}
    \label{fig:pipeline}
\end{figure*}

We now present the steps of our multiple-hypothesis path planning method, illustrated in Figure \ref{fig:pipeline}. Our planner takes as input the Gaussian obstacle estimates defined in section \ref{sec:obstacle_estimation}, constructs a \graphname{} that probabilistically represents possible routes through the forest, generates and evaluates multiple path hypotheses, and finally outputs a local goal point for the local planner described in section \ref{sec:local_planner}.

\subsection{Probability model for safe navigation between estimated obstacles}\label{sec:safety_prob}

\begin{figure}
    \centering
    \includegraphics[width=\columnwidth]{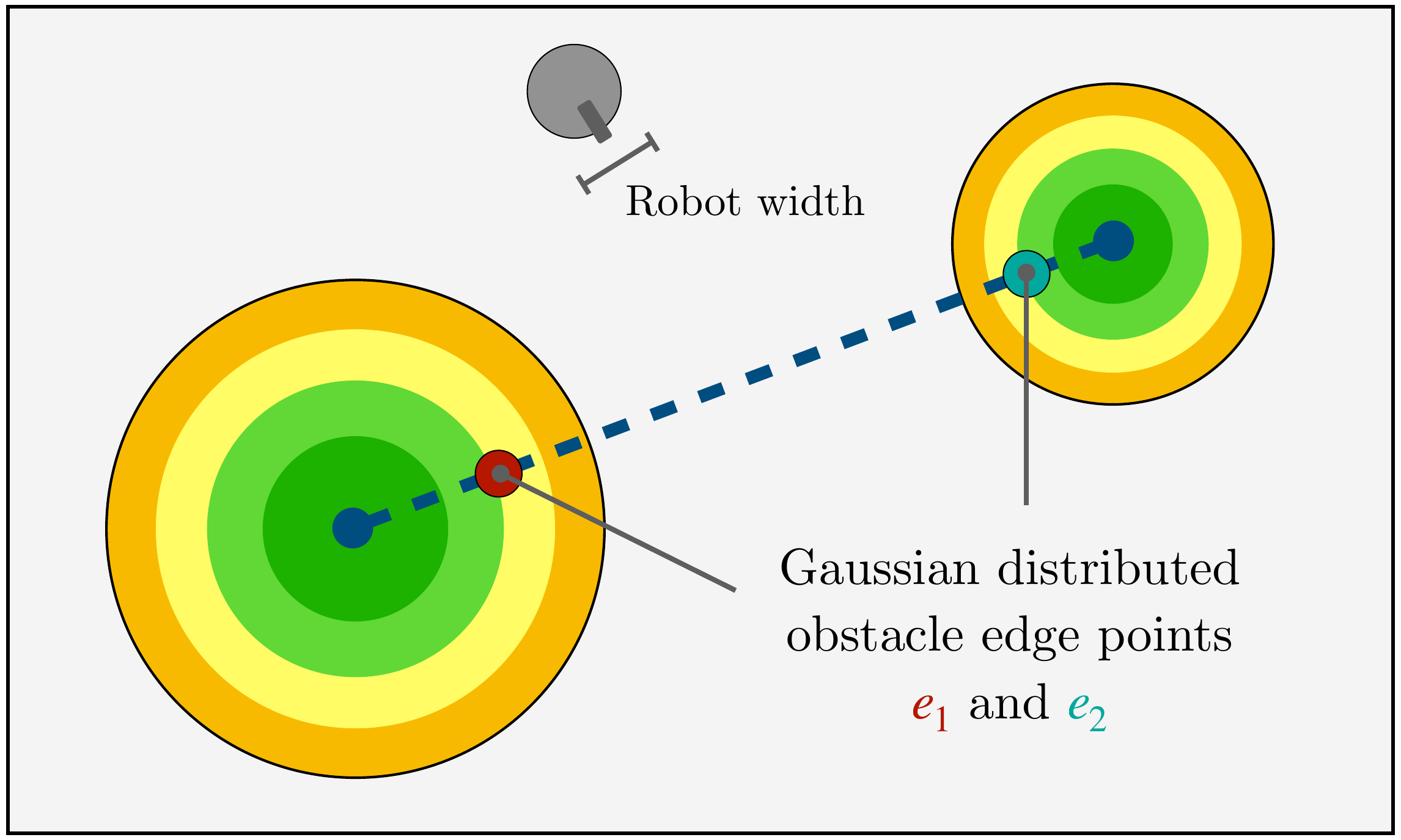}
    \caption{Our approach models the probability that a robot of a given width is able to safely pass between a pair of circular obstacles with Gaussian distributed uncertainty in their 2D positions and radii.}
    \label{fig:safety_prob}
\end{figure}

At a high level, a path through the forest is defined by a series of pairs of obstacles, between which the robot passes as it moves towards the goal. In order to reason about the safety of a given path through the forest, we derive a model for calculating the probability that the robot will be able to safely move between a pair of obstacles whose positions and sizes are estimated assuming Gaussian uncertainty, as defined in section \ref{sec:obstacle_estimation}. Figure \ref{fig:safety_prob} illustrates this problem.

\begin{figure}
    \centering
    \includegraphics[width=\columnwidth]{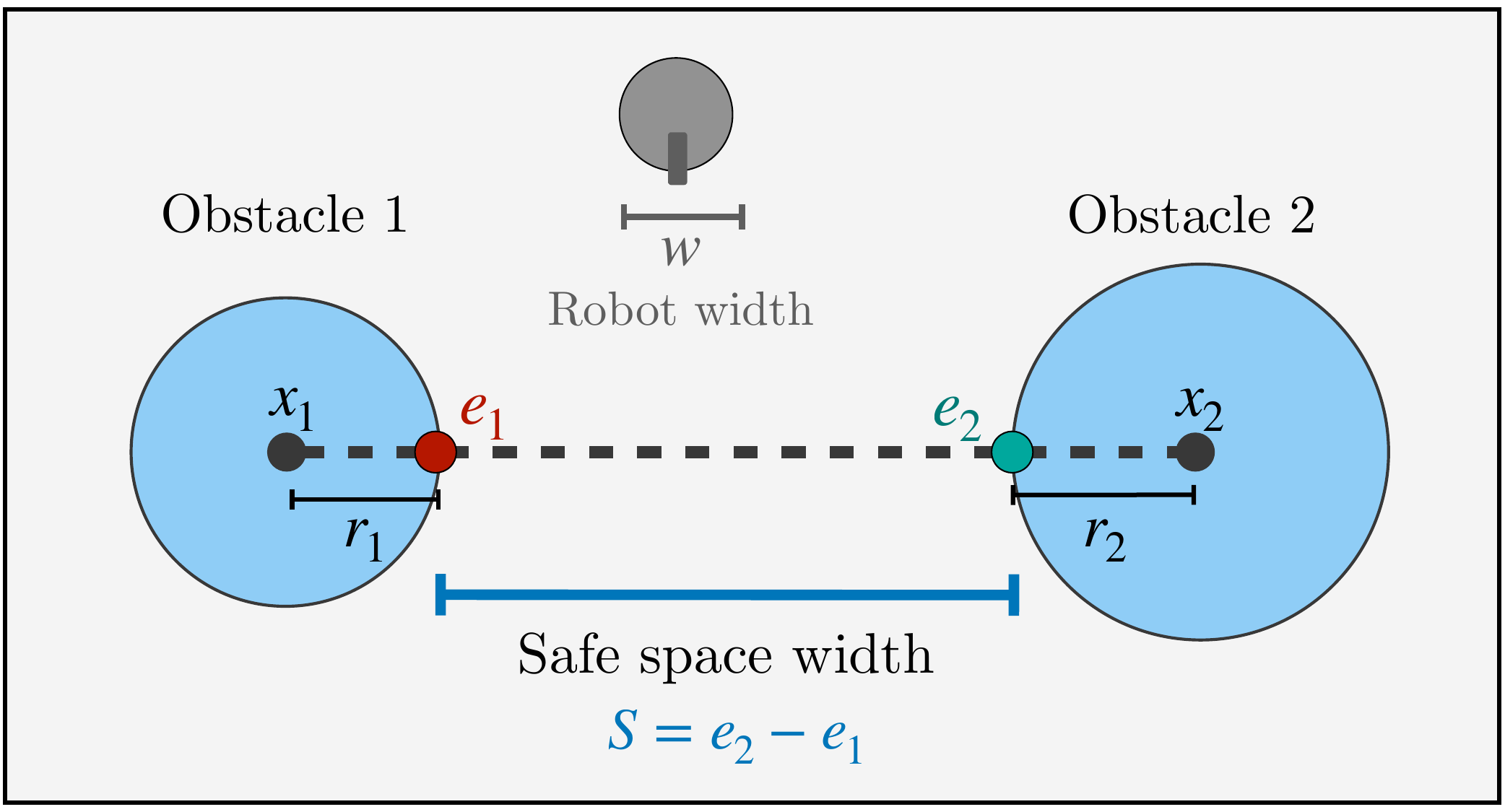}
    \caption{Illustration of the width of the safe space between a pair of obstacles in 1D, which is used to calculate the probability that the robot can safely pass between the obstacles.}
    \label{fig:safety_prob_1d}
\end{figure}

We begin by calculating the probability that a robot is able to safely move between a pair of one-dimensional obstacles, then generalize this simpler calculation to the two-dimensional case.

\subsubsection{1D safety probability calculation}

In one dimension, two obstacles $\obsgt_i$ and $\obsgt_j$ have scalar positions $x_i$ and $x_j$, and widths $d_i$ and $d_j$. Assuming that $x_i < x_j$, if the gap between the right edge of obstacle $\obsgt_i$ and the left edge of obstacle $\obsgt_j$ is greater than the robot width, then the robot will be able to safely move between the obstacles. For clarity in the following derivations, we use the obstacle radii $r_i = \frac{d_i}{2}$, $r_j = \frac{d_j}{2}$. The right edge point of $\obsgt_i$ is then $e_i = x_i + r_i$, and the left edge of $\obsgt_j$ is $e_j = x_j - r_j$. Then, the width of the free space between these obstacles is
\begin{align}\label{eq:freespace}
    S &= e_j - e_i\\
    & = x_j - x_i - (r_i + r_j).
    \end{align}

Figure \ref{fig:safety_prob_1d} shows an example illustration of the free space $S$, calculated based on the 1D positions and radii of a pair of obstacles.

Since we do not have perfect environment knowledge, we model the obstacle positions and sizes as Gaussian random variables, with the uncertain position and size of obstacle $\obs_i$ being
\begin{equation}
    \hat{x}_i \sim \mathcal{N} \left(\mu_{x,i}, \sigsq_{x,i}\right),
\end{equation}
\begin{equation}
    \hat{r}_i \sim \mathcal{N} \left(\mu_{r,i}, \sigsq_{r,i}\right),
\end{equation}
and likewise for $\obs_j$.

The estimated width of the free space $S$, as a linear combination of Gaussian random variables according to \eqref{eq:freespace}, is then also Gaussian distributed as
\begin{equation}\label{eq:S_gaussian}
    \hat{S} \sim \mathcal{N}\left(
            \mu_S, \sigsq_S
      \right),
\end{equation}
where
\begin{equation}\label{eq:freespacemean}
    \mu_{S} =  \left(\mu_{x,j} - \mu_{r,j}\right) - \left(\mu_{x,i} + \mu_{r,i}\right),
\end{equation}
\begin{equation}\label{eq:freespacestd}
    \sigsq_{S} =  \sigsq_{x,i} + \sigsq_{r,i} + \sigsq_{x,j} + \sigsq_{r,j}.
\end{equation}

Let the known width of the robot be $\wr$. Then, the probability that our robot can move between obstacles $\obs_i$ and $\obs_j$ is the probability that the space between them is enough to avoid collision, i.e. $P(S>\wr)$. We can exactly compute this probability as $1$ minus the cumulative distribution function (CDF) of $S$, giving us an expression for the safety probability using the Gaussian CDF equation:
\begin{equation}\label{eq:safetyprob1d}
    P(S > \wr) = \frac{1}{2}\left(1 - \text{erf}\left(\frac{x - \mu_S}{\sigma_S\ \sqrt{2}}\right)\right),
\end{equation}
where the mean and standard deviation of $S$, respectively $\mu_S$ and $\sigma_S$, are given by \eqref{eq:freespacemean} and \eqref{eq:freespacestd}. $\text{erf}$ denotes the error function used to compute the Gaussian CDF.

\subsubsection{2D safety probability calculation}

We now solve for the probability that the robot can safely navigate between a pair of 2D circular obstacles specified by their positions and diameters, showing that the 2D problem can be transformed to the 1D case without loss of generality.

Given a pair of 2D obstacle estimates $\obs_i=\left(\meanvec_{\obs,i}, \Sigma_{\obs,i} \right)$ and $\obs_j=\left(\meanvec_{\obs,j}, \Sigma_{\obs,j} \right)$, as defined in \eqref{eq:obs_estimate},
the angle between the world frame $x$-axis and the line passing through both position estimate means is 
\begin{equation}
    \theta_{ij} = \arctan\left(\frac{\mu_{y,j} - \mu_{y,i}}{\mu_{x,j} - \mu_{x,i}}\right).
\end{equation}

We can then compute the rotation matrix $\rot\in SO(2)$, which rotates the obstacle mean positions into a reference frame where the obstacle centers lie along the transformed x-axis,
\def\tij{\left(\theta_{ij}\right)}
\begin{equation}
    \rot = \begin{bmatrix}
    \cos{\tij} & \sin{\tij} \\
    -\sin{\tij} & \cos{\tij}
    \end{bmatrix}.
\end{equation}

Note that the negative $\sin$ term is located at the bottom-left of $\rot$, as this is a rotation by $-\theta_{ij}$.

We define the mean vector and covariance matrix which relate to only the position of obstacle $i$, omitting the obstacle diameter, as
\begin{equation}
    \meanvec_{xy,i} = \begin{bmatrix} \mu_{x,i} & \mu_{y,i} \end{bmatrix}^T
\end{equation}
\begin{equation}
    \Sigma_{xy, i} = \begin{bmatrix}
                    \sigma^{2}_{xx,i} & \sigma^{2}_{xy,i} \\
                    \sigma^{2}_{xy,i} & \sigma^{2}_{yy,i}
                            \end{bmatrix},
\end{equation}
where $\Sigma_{xy, i}$ is the upper-left $2\times2$ submatrix of the full covariance matrix $\Sigma_i$.

We transform the position mean and covariance using $\rot$, obtaining the transformed mean and transformed covariance matrix
\begin{equation}
    \meanvec_{xy, i}' = \rot\meanvec_{xy, i}
\end{equation}
\begin{equation}
    \Sigma_{xy, i}' = \rot\Sigma_{xy, i} \rot^T,
\end{equation}
and likewise for the transformed mean and covariance of obstacle $\obs_j$.

We then marginalize over the transformed $y$-axis, to find the position mean and variance along the transformed $x$ axis. This leave us with the simpler 1D probability calculation, and we can then apply the equations for the 1-D safe navigation probability, along the transformed $x$-axis. 

Defining the components of the transformed mean and covariance of obstacle $i$ as
\begin{equation}
    \meanvec_{xy, i}' = \begin{bmatrix} \mu'_{x,i}  & \mu'_{y,i}\end{bmatrix}^T
\end{equation}
\begin{equation}
    \Sigma_{xy, i}' = \begin{bmatrix}
                    \sigma^{2'}_{xx,i} & \sigma^{2'}_{xy,i} \\
                    \sigma^{2'}_{xy,i} & \sigma^{2'}_{yy,i}
                            \end{bmatrix},
\end{equation}
we can then express the mean and variance of the free space $S$ between the two obstacles, \emph{in the case of two-dimensional circular obstacles}, as
\begin{equation}\label{eq:freespacemean2d}
    \mu_{S} =  \left(\mu'_{x,j} - \mu_{r,j}\right) - \left(\mu'_{x,i} + \mu_{r,i}\right),
\end{equation}
\begin{equation}\label{eq:freespacestd2d}
    \sigsq_{S} =  \sigma^{2'}_{xx,i} + \sigma^{2}_{r,i} + \sigma^{2'}_{xx,j} + \sigsq_{r,j}.
\end{equation}

Using these terms, we can then compute the probability of safe navigation between the two obstacles using equation \eqref{eq:safetyprob1d}.

%% file: approach_subsections/graph-construction.tex

\subsection{Navigation graph construction}

\begin{figure*}
    \centering
    \includegraphics[width=\textwidth]{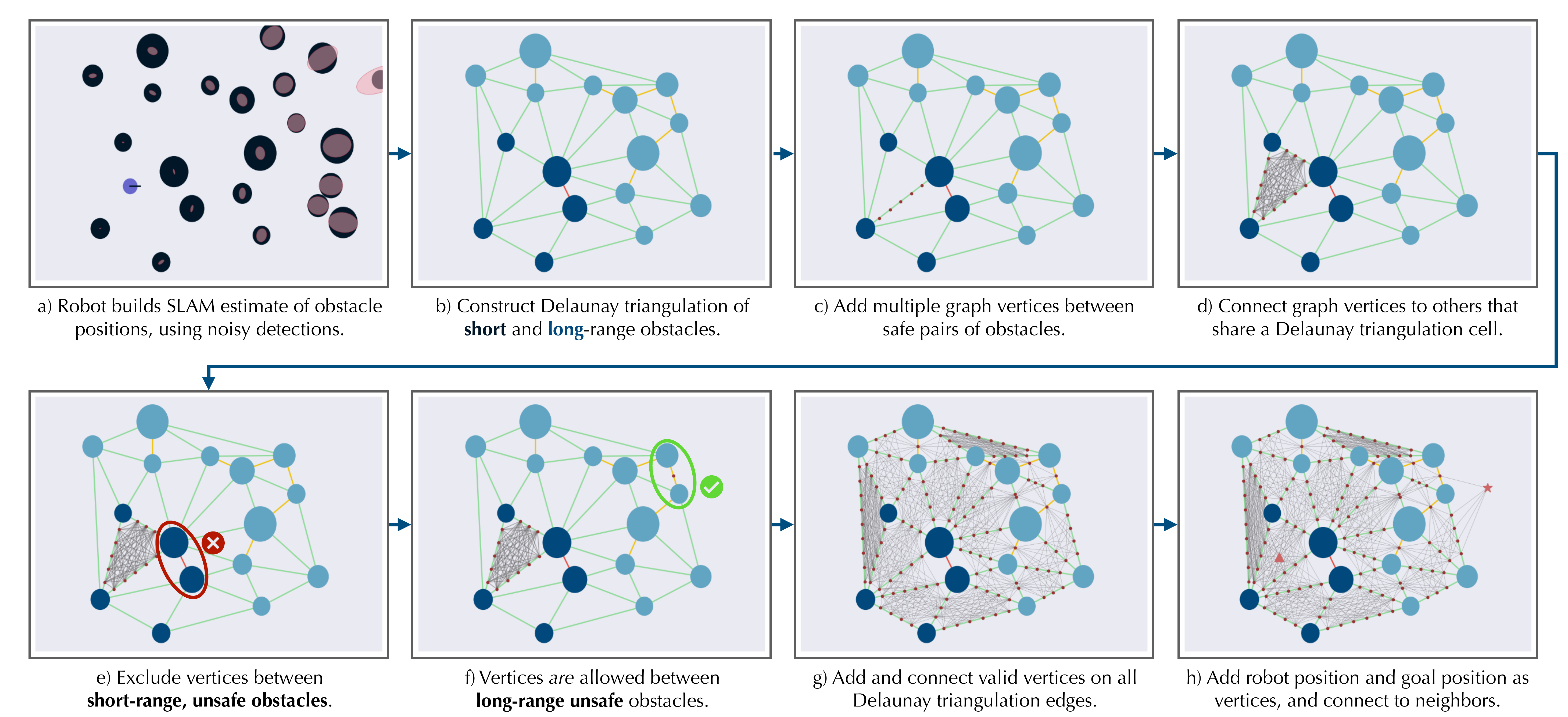}
    \caption{Steps of the graph construction process.}
    \label{fig:graph_construction}
\end{figure*}

Given estimates of the obstacle positions and sizes, we construct a graph which represents the high-level paths the robot can take through the environment to reach its goal. Our graph encodes information on \emph{path distance} and \emph{path safety}, while also taking into account the \emph{uncertainty in obstacle estimates}. Figure \ref{fig:graph_construction} illustrates the graph construction, and Algorithm \ref{alg:graph_construction} lists the steps of the process. 

\begin{algorithm}
\caption{Constructing the \graphname{}}
\label{alg:graph_construction}
\textbf{Required:} A method for computing the probability that the robot can safely pass between a pair of obstacles, $P_{safe}(\obs_i, \obs_j)$, which returns a probability in the set $\{p \in \mathcal{R} \mid 0 \leq p \leq 1\}$.
\begin{algorithmic}[1]
\Require
    \Statex $\obs_i = \left(\vec{\mu}_{O,i}, \Sigma_{O, i}\right), i\in[1, \nobs]$, obstacle estimates.
    \Statex $\start$,    the robot's starting state.
    \Statex $\globalgoal$,    the global goal position.
    \Statex $\desiredprob\in[0,1]$, the desired safety probability.
    \Statex $\rangeshort \in \nonneg$, the threshold between the short and long range zones.
    \Statex $w_{robot} \in \nonneg$, the robot width.
\Ensure 
    \Statex $\vnav, \enav$, the \graphname{} structure.
    \Statex $\costdist$,    the edge distance costs.
    \Statex $\probsafe$,    the vertex safety probabilities.
    \Statex $\range$,       the vertex range zones.
\State $\vdt, \edt = DT(\allobs)$ \Comment{Delaunay triangulation.}
\State $\vnav \gets \emptyset$ \Comment{Initialize \graphname{} vertices.}
\State $\enav \gets \emptyset$ \Comment{Initialize \graphname{} edges.}
\ForAll{$e_{DT, ij}\in\edt$} \Comment{Delaunay edge between obstacles i, j.}
    \LineComment{Compute distance from robot to obstacles.}
    \State $r_i = compute\_distance(\vec{\mu}_{O,i}, \start)$
    \State $r_j = compute\_distance(\vec{\mu}_{O,j}, \start)$
    \If{$r_i > \rangeshort$ \textbf{or} $r_j > \rangeshort$}
        \State $r \gets long$
    \Else
        \State $r \gets short$
    \EndIf
    \State $prob = P_{safe} (\obs_i, \obs_j)$
    \If{$prob < \desiredprob$}
        \If{$r = short$}
            \State $V_{new} = \emptyset$ 
        \Else
            \State $V_{new} = \{midpoint(\meanvec_{O,i}, \meanvec_{O,j})\}$
        \EndIf
    \Else
        \State $V_{new} = \{place\_points(\obs_i, \obs_j, w_{robot})\}$
    \EndIf
    \ForAll{$v_k \in V_{new}$}
        \State $\probsafe(v_k) = prob$
        \State $\costsafe(v_k) = -\log(prob)$
        \State $\range(v_k) = r$
        \ForAll{$v_l \in \vnav$ that share a Delaunay cell with $v_k$}
            \State $\enav \gets \enav \cup \{e_{kl}\}$
            \State $\costdist(e_{kl}) = distance(v_k, v_l)$
        \EndFor
    \EndFor
    \State $\vnav \gets \vnav \cup V_{new}$
\EndFor
\State \Return $(\vnav, \enav, \costdist, \probsafe, \costsafe, \range)$
\end{algorithmic}
\end{algorithm}

\subsubsection{Delaunay triangulation cell decomposition}

Previous works have used the Delaunay triangulation for localization in forests, due to its ability to qualitatively describe regions in the environment\cite{li2020localization, nardari2020place}. Using the Delaunay triangulation algorithm \cite{lee1986delaunay}, we divide the workspace into triangular cells with obstacle center points as their vertices. The Delaunay triangulation computes these cells such that each triangular cell, defined by a triplet of obstacles, contains no other obstacle center points within its boundaries. Figure \ref{fig:graph_construction}.b) shows an example Delaunay triangulation.

The edges between the Delaunay triangulation cells, which we will refer to as \emph{cell faces} (to avoid confusion with the navigation graph edges which we will define later), represent thresholds where the robot passes \emph{between} a pair of obstacles, as described in \cite{shah2013qualitative}. 

The Delaunay triangulation is defined for our purposes as $DT(\allobs) = (\vdt, \edt, \tdt)$, where $\allobs$ is the set of estimated obstacles. $\vdt$ are the vertices of the triangulation, placed at the mean center positions of the estimated obstacles, so that $\vdt = \{\meanvec_{xy,i} \mid \obs_i \in \allobs\}$. $\edt$ is the set of Delaunay triangulation cell faces, while $\tdt$ is the set of obstacle triplets that form the cells; therefore, $\exists (\obs_i, \obs_j, \obs_k) \in \tdt \iff \exists e_{ij}, e_{jk}, e_{ik} \in \edt$. The circle circumscribed around the vertices of the triangle formed by these obstacles contains no other obstacle center points $\meanvec_{\obs, l}$ such that $l \notin \{i, j, k\}$.

For each Delaunay cell face between a pair of obstacles $\obs_i$ and $\obs_j$, we use Equation \ref{eq:safetyprob1d} to calculate the probability that the robot can safely pass between $\obs_i$ and $\obs_j$, transitioning from one Delaunay cell to the next. Since at a high level, a path through the forest can be thought of as a series of obstacle pairs between which the robot should pass, we can therefore equivalently define a path by a series of Delaunay triangulation cells. In the following sections of this paper, we define a graph structure over the Delaunay cells, over which we can search in order to determine possible paths to the goal.

\subsubsection{Definitions of graph terms}\label{sec:graph_constructions_defs}

Before introducing our \graphname{} construction procedure, we define the following terms:

\begin{itemize}
    \item $\desiredprob \in [0,1)$, the \emph{desired safety probability}, specified as a planner parameter. The planner will attempt to find a path whose safety (i.e. the probability that the robot will be able to reach the global goal without crashing into an obstacle by following this path) exceeds this threshold.
    \item $\gnav = \left(\vnav, \enav\right)$, the \textbf{navigation graph{}}, defined by its vertices and edges. The vertices of this graph are placed on the Delaunay triangulation cell faces, and are connected by edges to other vertices on adjacent cell faces. A search over this graph therefore finds a path passing between a series of obstacle pairs in the forest. Figure \ref{fig:graph_construction}.h) illustrates an example \graphname{}, showing the placement of vertices and edges.
    \item $v_i \in \vnav, e_{ij} \in \enav$, the individual vertices and edges.
    \item $\costdist: e_{ij} \rightarrow \nonneg$, a mapping which gives the Euclidean distance cost for each navigation graph edge. $\nonneg$ is the set of non-negative real numbers.
    \item $\probsafe: v_{i} \rightarrow \{p \in \mathbb{R} \mid 0 \leq p \leq 1\}, \forall i \in \vnav$,  a mapping which gives the safety probability for each graph vertex, i.e. the probability that the robot can safely move between the pair of obstacles associated with this vertex.
    \item $\costsafe: v_{i} \rightarrow \nonneg,$ defined as the safety cost $\costsafe(v_i) = -\log(\probsafe(v_i)),\forall i \in \vnav$. 
    \item $\setofranges=\{short, long\}$, the set of \textbf{range zones}. Our planner treats vertices differently depending on their range zone, due to differing amounts of sensor noise and estimation uncertainty at different sensor ranges.
    \item $\rangeshort \in \nonneg$, the range threshold defining the two range zones.  The threshold is defined as a planner parameter. 
    \item $\range: v_i \rightarrow \setofranges$, a \textbf{range mapping} which maps a vertex to the range zone to which it belongs.
\end{itemize}

\subsubsection{Vertex range zones} 

Obstacle estimation uncertainty varies significantly with distance from the robot, due to increasing sensor measurement noise with range, and our graph structure therefore stores obstacle range information so that the planner can account for this variation.

As illustrated in Figure \ref{fig:graph_construction}.b), we divide the estimated obstacles into two \emph{range zones}: short and long, divided by the threshold $\rangeshort$. Obstacles at distance less than $\rangeshort$ are considered short range, and all others are considered long range. We assume that at short range, obstacle position and size estimates are fairly confident, having been updated recursively with multiple range-bearing measurements from detections. Beyond the short range, obstacle estimates contain significant uncertainty due to higher measurement noise at further ranges. 

\subsubsection{Navigation graph vertex placement}

The placement of vertices in the graph depends on obstacle estimate safety probabilities as well as their range zones. For safe Delaunay cell faces, i.e. $\probsafe(e_{DT,ij}) \geq \desiredprob$, we allow the robot to plan paths between the estimated obstacles $\obs_i$ and $\obs_j$, and we add corresponding vertices to the graph as shown in Figure \ref{fig:graph_construction}.c). 

For an unsafe Delaunay cell face, i.e. $\probsafe(e_{DT,ij}) < \desiredprob$, the decision on whether or not to add graph vertices along this cell face depends on the range zone of obstacles $\obs_i$ and $\obs_j$. At short range, we do not place graph vertices on unsafe Delaunay cell faces, as demonstrated in Figure \ref{fig:graph_construction}.e). When planning around close-by obstacles, the robot may not have enough time to collect more measurements and update its estimates of these obstacles. Therefore, the robot should not risk a crash by passing between unsafe obstacle pairs in the short range.

In contrast, path planning between uncertain long-range obstacles should take into account the fact that the robot will have time to safely collect additional sensor information, and update its estimate of these obstacles before closely approaching them. Preventing the robot from planning between uncertain, far-off obstacles will likely lead to excessively conservative planning behavior in obstacle-dense environments. During the graph construction, we therefore place \graphname{} vertices on Delaunay cell faces between unsafe long-range obstacles, recognizing that as the robot receives more detections, some of these uncertain vertices will resolve to safe and viable navigation paths. The multiple-hypothesis planner we introduce in the next section of this paper intelligently considers multiple of these candidate paths, making use of uncertain information from the obstacle estimates. 

We place multiple vertices on long Delaunay cell faces which are determined to be very safe. These cell faces correspond to very wide gaps between obstacles, where the overall path distance may change significantly depending on whether the robot decides to pass nearby one of the obstacles or the other. Placing multiple vertices on these cell faces therefore allows the graph search to better approximate the geometric length of paths.

For each new vertex added on Delaunay cell face $e_{DT, ij}$, we assign the vertex to the short range zone if both obstacles $\obs_i$ and $\obs_j$ are within the short range threshold $\rangeshort$, and assign the vertex to the long range zone if \emph{either} obstacle is outside the short range.

We record the Delaunay cell face which each graph vertex lies on, and we additionally save the safety probability of the Delaunay cell face as $\probsafe(v_k) = P_{safe}(\obs_i, \obs_j), \forall k \in V_{DT}$. The saved range zone and safety probability information will later be used by the multiple-hypothesis planner.

\subsubsection{Graph edge construction}

For each Delaunay cell, we add graph edges to $\enav$ connecting each vertex on one of the cell faces with all vertices on other cell faces, as shown in Figure \ref{alg:graph_construction}.d). For each $e_{ij} \in \enav$, we set the distance cost $\costdist(e_{ij})$ to the 2D Euclidean distance between the 2D positions of vertices $v_i$ and $v_j$.

Finally, we add the robot start position and goal positions as graph vertices $v_{start}$ and $v_{goal}$, shown in Figure \ref{alg:graph_construction}.h). If the start and/or goal lie within a Delaunay triangulation cell, we connect them via graph edges, weighted by 2D Euclidean distance, to all graph vertices that lie on the faces of this cell. If either the start or goal lies outside all Delaunay cells, we connect them to all visible vertices on the cell faces at the boundary of the Delaunay triangulation (i.e. all vertices which can be connected by a straight line segment in 2D to the start/goal, without the line segment passing through any other cell faces). We set $\probsafe(v_{start}) = \probsafe(v_{goal}) = 1.0$, i.e. the robot does not incur any safety cost traveling through the start and goal vertices.

%% file: approach_subsections/graph-planning.tex

\subsection{Multiple-hypothesis planning under uncertainty}

\begin{figure*}
    \centering
    \includegraphics[width=\textwidth]{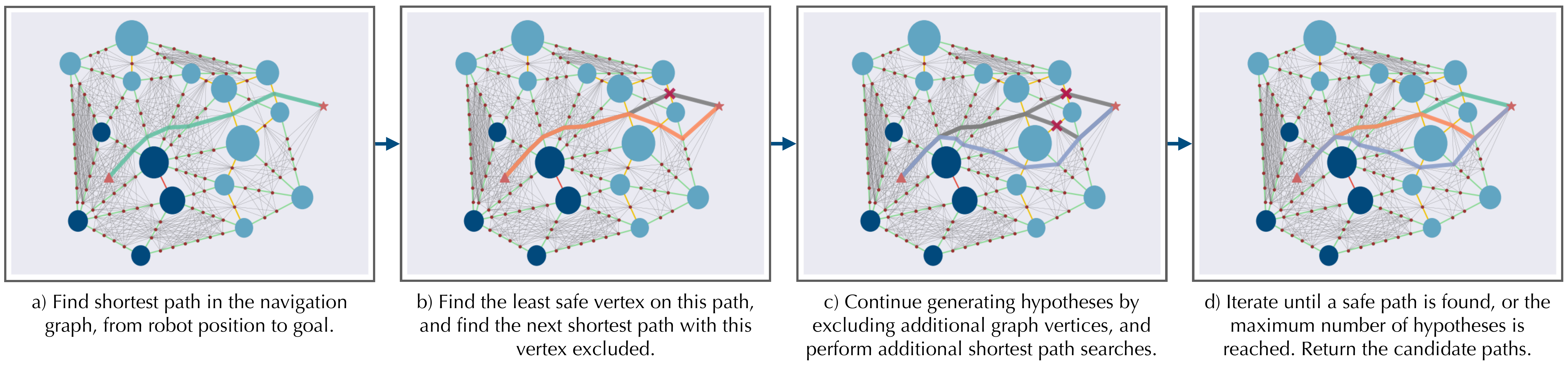}
    \caption{Example of the multiple-hypothesis planning procedure using the \graphname{}. The planner generates hypotheses of the environment, until finding a safe path--in this case, the pink path--with a safety probability over the desired safety $\desiredprob$, or until generating the maximum allowed number of hypotheses.}
    \label{fig:graph_planning}
\end{figure*}

We now present a high-level planning method which uses the constructed \graphname{} to generate  and evaluate candidate paths to the goal, based on multiple environment hypotheses that take into account the uncertainty in obstacle estimates. Algorithm \ref{alg:graph_planning} outlines the multiple-hypothesis planning algorithm.

\subsubsection{Definitions of planner terms}

We define the following terms, used in addition to those defined in \ref{sec:graph_constructions_defs} to describe our multiple hypothesis planning algorithm.

\begin{itemize}
    \item $\hyp: v_i \rightarrow \{0, 1\}$, a \textbf{hypothesis} of the environment, defined as a mapping which takes as input a \graphname{} vertex and outputs either a $0$, indicating that \emph{in this hypothesis}, the vertex is not safe to pass through, or a $1$, indicating that it is safe.
    \item $\path \subset \vnav$, a \textbf{path} through the graph, defined as an \emph{ordered} subset of the graph vertices. For any vertex $v_i$ at index $\path[k]$ in the path, and vertex $v_j$ at index $\path[k+1]$, there must exist a graph edge $e_{i,j}\in \enav$.
\end{itemize}

The planner uses the following parameters:

\begin{itemize}
    \item $\nhyp \in \mathbb{N}_{\textbf{}>0}$, the maximum number of path hypotheses to generate.
    \item $\minprob \in [0,1)$, with $\minprob < \desiredprob$ a minimum safety probability for a graph vertex to be considered by the multiple hypothesis planner. This parameter is optional, and by default can be set to zero, causing it to have no effect on the planner.
    \item $\wdist \in \nonneg$, the weight on path length, and $\wsafe \in \nonneg$, the weight on path safety. These weights are used to evaluate the candidate paths in order to decide on the best overall path.
\end{itemize}

\subsubsection{Hypothesis generation}

Our planner requires a shortest path search algorithm $SP(\vnav, \enav, \costdist, \hyp)$. The input $\hyp$ is a \emph{hypothesis}, which excludes some subset of the vertices $v_i \in \vnav$ that are considered to be unsafe (i.e. $\hyp(v_i) = 0$).

We first construct an initial hypothesis $\hyp_0$ with all graph vertices marked as safe, i.e. $\forall v_i, \hyp_0(v_i) = 1$. Optionally, if a minimum probability $\minprob$ greater than zero is specified, for any vertices with $\probsafe(v_i) < \minprob$, we set $\hyp_0(v_i)$ to 0. The threshold $\minprob$ can be used to exclude extremely unsafe vertices from the multiple hypothesis planner, making the path search more efficient.

The planner then computes the shortest path $\path_0 = SP(\vnav, \enav, \costdist, \hyp_0)$, finding the shortest possible path to the goal according only to the Euclidean distances given by the \graphname{} edge distance costs $\costdist(e_{ij})$. We initialize a set of candidate paths $\paths$, at first containing only $\path_0$. Figure \ref{fig:graph_planning}.a) illustrates this step.

The planner then generates additional candidate paths, up to $\nhyp$ in total, each planned under a different hypothesis of which vertices in the graph are safe to pass through. We store the graph vertices $v_i \in \path_0$ in a priority queue $\queue$. Each vertex is stored along with a copy of $H_0$. We denote the priorities used for the queue as $\pi_{0,i}$ and calculate the priority for each vertex $v_i$ as $\pi_{0,i} = -(1-\probsafe(v_i))$, the negative of the likelihood that vertex $v_i$ is unsafe. Computing the priorities in this way allows us to draw the vertex which is most likely to be unsafe from the queue at each iteration of the multiple hypothesis path search.

At each planner iteration, we use $\queue$ to find the graph vertex $v_{i}$ which is most likely to be unsafe. We pop this vertex from the queue, with its associated hypothesis $\hyp_j$, then initialize a new hypothesis $\hyp_k$ as a copy of $\hyp_j$. We then set $\hyp_k(v_i) = 0$. 

We then plan the shortest path \emph{under this hypothesis}, $\path_k = SP(\vnav, \enav, \costdist, \hyp_k)$, finding the shortest path assuming that the graph vertex $v_i$ is unsafe. This new path will be added to the set of candidate paths if it is not a duplicate of an existing path, and if the safety of the path evaluated only over the short range vertices in the path is above $\desiredprob$. This condition guarantees that the robot will not execute a path that is unacceptably unsafe in the short range.

If the new path $\path_k$ passes these checks, we add it to the set of candidate paths $\paths$, and add all vertices on the path to the queue $\queue$, along with hypothesis $\hyp_k$. For each vertex $v_l \in \path_k$, the priority $\priority_{k,l}$ is stored as $(1 - \probsafe(v_l)) \cdot \priority_{k,i}$, indicating that the likelihood of a hypothesis depends conditionally on multiple graph vertices being unsafe.  Figure \ref{fig:graph_planning}.b-c show the iterative process of planning, excluding graph vertices using the priority queue, then planning again under the new hypothesis.

We terminate the search once the maximum number of hypotheses has been reached, i.e. $\vert \paths \vert = \nhyp$, or a path $\path_i$ is found such that the safety of $\path_i$ is above the desired safety probability, i.e. $\prod_{v_j \in \path_i} \probsafe(v_j) \geq \desiredprob$. In calculating the path safety in this way, we assume that the values $\probsafe(v_i)$ of different graph vertices are independent of one another. The output of the search process is a set of candidate paths through the \graphname{}, as shown in the example in Figure \ref{fig:graph_planning}.d.

\subsubsection{Candidate path evaluation}\label{sec:path_eval}

Given the set of paths $\paths$, we calculate the \emph{distance} and \emph{safety} costs for each path, as
\begin{equation}
    \pathcostdist(\path_i) = \sum_{(v_j, v_k) \in \path_i} \costdist(e_{jk}),
\end{equation}
and
\begin{equation}
    \pathcostsafe(\path_i) = \sum_{v_j \in \path_i} \costsafe(v_j).
\end{equation}

Since the distance costs and safety costs are on different scales, we normalize both across all paths. For all $\path_i \in \paths$:
\begin{equation}
    \pathcostdist(\path_i) = \frac{\pathcostdist(\path_i)}{\max_{\path_j \in \paths} \pathcostdist(\path_j)},
\end{equation}
\begin{equation}
    \pathcostsafe(\path_i) = \frac{\pathcostsafe(\path_i)}{\max_{\path_j \in \paths} \pathcostsafe(\path_j)}.
\end{equation}

Finally, we compute the overall cost of each path
\begin{equation}
    \pathcosttotal(\path_i) = \wdist \cdot \pathcostdist(\path_i) + \wsafe \cdot \pathcostsafe(\path_i),
\end{equation}
and use the final path with minimum cost
\begin{equation}
    \bestpath = \argmin_{\path_i \in \paths} \left(\pathcosttotal(\path_i)\right).
\end{equation}

\subsubsection{Local goal computation}\label{sec:local_goal}

Starting from the robot position, at vertex $v_0$ in $\bestpath$, we then calculate the 2D goal point for the local planner, $\localgoalxy$, by finding the point along the 2D line segments $e_{ij}$, such that $\left(v_i, v_j\right) \in \bestpath$, which is ahead of the robot by a plan-ahead distance $\planaheaddist \in \nonneg$. This approach uses the high-level path $\bestpath$ to guide the local planner, which handles the details of generating a dynamically-feasible trajectory for the robot and accounting for the specific geometry of nearby obstacles.

\begin{algorithm}
\caption{Multiple-hypothesis planning in the \graphname{}}
\label{alg:graph_planning}
\textbf{Required:} A shortest path graph search algorithm $SP(\vnav, \enav, \costdist, \hyp)$ which returns a path $\path$.
\begin{algorithmic}[1]
\Require
    \Statex $\vnav, \enav$, the graph structure.
    \Statex $\costdist$,    the edge distance costs.
    \Statex $\probsafe$,    the vertex safety probabilities.
    \Statex $\range$,       the vertex range zones.
    \Statex $\desiredprob$, the desired safety probability.
    \Statex $\minprob$, the minimum vertex safety to consider.
\Ensure 
    \Statex $\paths$, a set of candidate paths.
\Procedure{Generate initial hypothesis $\hyp_0$}{}
\ForAll{$v_i\in\vnav$}
\If{$\probsafe(v_i) < \minprob$}
    \State$\hyp_0(v_i) \gets 0$ \Comment{0 indicates unsafe vertex.}
\Else
    \State{$\hyp_0(v_i) \gets 1$} \Comment{1 indicates safe vertex.}
\EndIf
\EndFor
\EndProcedure
\State Find the shortest path $P_0 = SP(\vnav, \enav, \costdist, \hyp_0)$.
\State Initialize empty priority queue $\queue$.
\ForAll{$v_i \in P_0$}
    \LineComment{Use vertex safeties to compute queue priorities.}
    \State $\priority_{0,i} \gets -(1 - \probsafe(v_i))$
    \State $queue\_insert\left(\queue, \priority_{0,i}, \left(v_i, H_0\right)\right)$
\EndFor
\State $\paths \gets \{\path_0\}$ \Comment{Initialize set of candidate paths.}
\Procedure{Multiple Hypothesis Search}{}
    \While{$\vert \paths \vert < \nhyp$}
        \LineComment{Get most unsafe vertex from $\queue$.}
        \State $\priority_{j,i}, v_i, \hyp_j = queue\_pop(\queue)$. 
        \State $\hyp_k \gets \hyp_j$ \Comment{Copy previous hypothesis.}
        \State $\hyp_k(v_i)\gets 0$  \Comment{Set new vertex to unsafe.}
        \State $P_k = SP(\vnav, \enav, \costdist, \hyp_k)$ \Comment{Path search.}
        \State $P_{k,short} = \{v_l \in P_k \mid R(v_l) = short\}$
        \If{$safety(P_{k,short}) < \desiredprob$}
            \State \textbf{continue} \Comment{Path unsafe in short range}
        \EndIf
        \If{$P_k \notin \paths$} \Comment{Disallow duplicate paths.}
            \State $\paths \gets \paths \cup \{P_k\}.$
            \If{$safety(P_k) \geq \desiredprob$}
                \State \textbf{break} \Comment{Found safe path, end search.}
            \EndIf
            \ForAll{$v_l \in P_k$}
                \State $\priority_{k,l} \gets (1-\probsafe(v_l)) \cdot \priority_{j,i}$
                \State $queue\_insert(\queue, \priority_{j,i}, (v_l, \hyp_k))$
            \EndFor
        \EndIf
    \EndWhile
\EndProcedure
\State \Return $\paths$
\end{algorithmic}
\end{algorithm}

%% file: approach_subsections/guarantees.tex

\subsection{Planner safety guarantees}

Given the graph structure, paths through the graph, and formal probabilities of a robot passing safely between a pair of obstacles, one can calculate probability guarantees for each path. 

Given a total number of vertices in $\path_i$, assuming each passing through a pair of obstacles is an independent event, the probability of a collision for $\path_i$ is
\begin{equation}
p_{\textrm{col}}(\path_i) = \left(1 -  \prod_{v_j \in \path_i} \probsafe(v_j) \right)
\end{equation}

One could also simply consider the worst case probability, or 
\begin{equation}
p_{\textrm{col}}(\path_i) \geq \left(1 - \min_{v_j \in \path_i} \probsafe(v_j) \right)
\end{equation}

%% file: experiments.tex


\input{experiments_subsections/simulations}
\input{experiments_subsections/qualitative}


%% file: experiments_subsections/simulations.tex

\section{Experiments and results}

\subsection{Simulated forest environments}\label{sec:sim_forests}

We evaluate our planner in a large-scale experimental trial, using randomly generated simulated forest environments. We vary the complexity of the simulated forests, in terms of the overall density of trees and their distribution, in order to test our hypothesis that our probabilistic planner should enable safer navigation in more complex and obstacle-dense environments.

\begin{figure}
    \centering
    \includegraphics[width=\columnwidth]{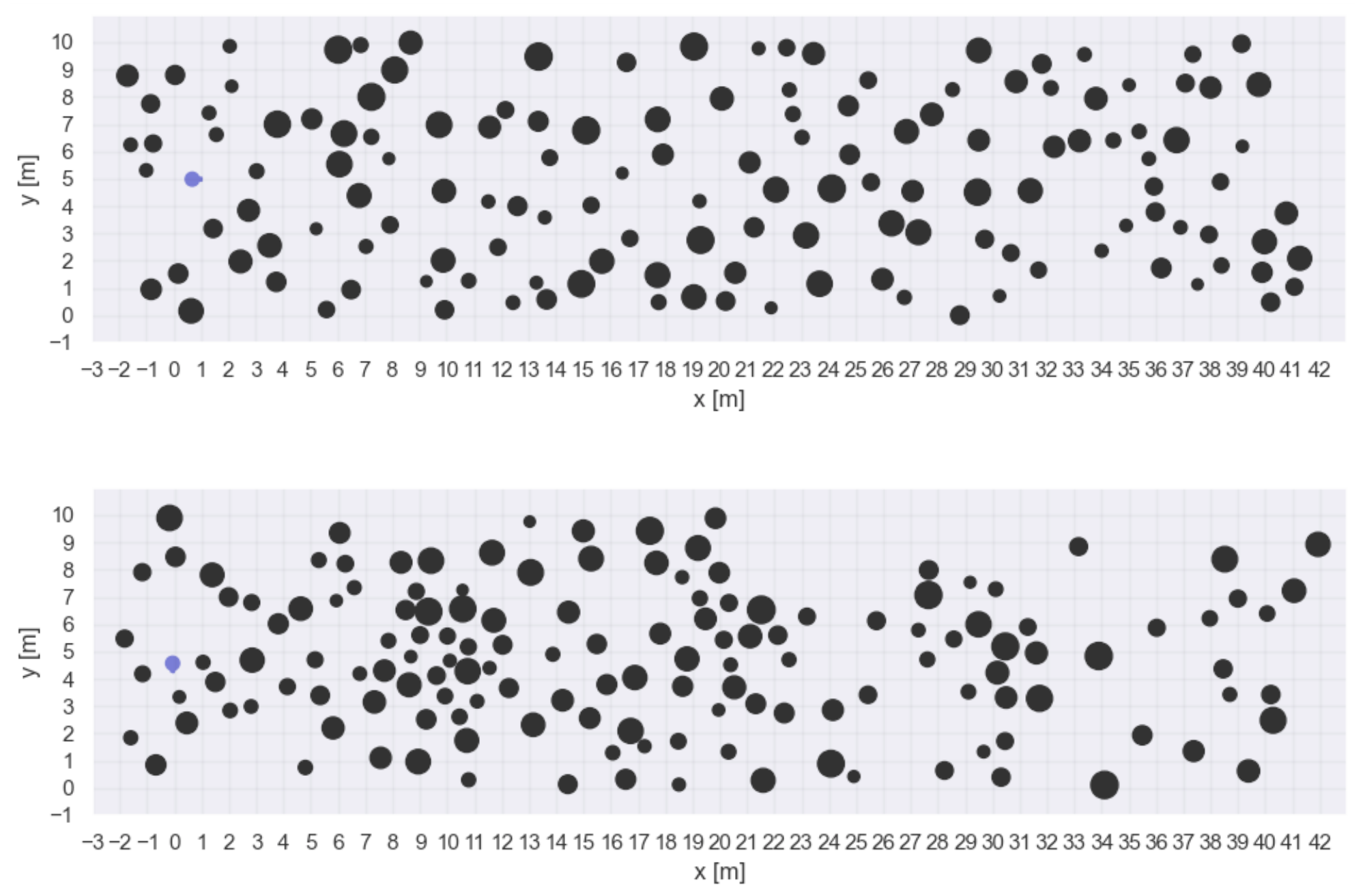}
    \caption{Example randomly generated forest environments, at the highest density setting $\rho = 0.3$ trees per square meter. The robot's goal is to navigate from $(x=0\si{m}, y=5\si{m})$ to $(x=40\si{m}, y=5\si{m})$. \textbf{Top:} Uniformly distributed tree positions. \textbf{Bottom:} Poisson forest with three Gaussian-distributed tree clusters added.}
    \label{fig:sim_forests}
\end{figure}

We generate our forests by specifying a desired tree density per area, then sampling the total number of trees in the environment according to the Poisson distribution as in \cite{karaman2012forestflight}. Tree sizes are uniformly distributed in a defined range, and we sample trees such that they do not overlap with each other. We first conduct simulations in forest environments of varying densities, where tree positions are uniformly sampled across the forest bounds. Noting that through areas of uniformly distributed trees, the best path through the forest mostly involves driving in a straight line to the goal with only small deviations, we then conduct an additional experiment in the more challenging case of forests containing three Gaussian-distributed clusters of trees (along with additional uniformly distributed obstacles), which lie directly between the robot and its goal. The robot generally cannot safely pass through these regions, so identifying and avoiding them preemptively is the most efficient way to reach the goal. Figure \ref{fig:sim_forests} shows two example simulated forests.

\subsection{Path planner system implementation}

\subsubsection{Obstacle estimation}

We define a detection range noise model consistent with the noise model for the Stereolabs ZED camera\footnote{https://support.stereolabs.com/hc/en-us/articles/206953039-How-does-the-ZED-work}. The standard deviation of the range measurement noise is proportional to the distance from the sensor, with this proportion growing quadratically with distance. The robot receives range and bearing measurements of all obstacles within a given maximum range and front-centered field of view, except for any obstacles which are fully occluded.

In order to estimate the obstacle positions, we perform factor graph SLAM with range and bearing obstacle measurements, as described in section \ref{sec:obstacle_estimation}. We use the iSAM2 \cite{kaess2012isam2} implementation provided by the GTSAM library\footnote{https://gtsam.org/}, and treat the obstacles as landmarks in landmark SLAM. The range-bearing measurements define binary (i.e. involving two variables in the graph) factors that probabilistically constrain the obstacle positions with respect to the robot pose at each time step.

We assume the position and size of an obstacle are decoupled, so the cross-covariance terms for position and size in $\obssigma$ are zero. We estimate the landmark sizes using simple 1D Kalman filters, in parallel with the SLAM obstacle position estimation.

In our simulation, the robot receives obstacle detections at a lower rate than odometry measurements. At time steps where no detections are available, we perform SLAM updates using odometry only. In order to isolate our study of the path planner from localization error effects, our experiments use simulated odometry measurements with negligible noise, so the localization problem is effectively solved and the estimation practically reduces to a mapping problem.

The true assignment of detections to landmark identities is unknown to the robot. Following \cite{kaess2008isam}, we perform data association using the Mahalanobis distance between new detections and existing landmark estimates, in order to account for the estimate uncertainties during assignment. In order to compute the Mahalanobis distance, we transform the obstacle estimate 2D position uncertainty to the range-bearing space. We discard any data association matches whose Mahalanobis distance falls over a set threshold, and initialize new landmarks for any detections that remain unmatched after association.

\subsubsection{Hybrid A* local planner}

We simulate a 2D differential drive robot which plans iteratively through the 2D forest environment. The robot uses a high-level global planner to specify a short-term goal for a local planner to follow, as described in section \ref{sec:local_planner}, and re-plans as it moves and obtains new sensor measurements.

For our experiments, we implement a hybrid A* local planner based on \cite{dolgov2010pathplanning}, \cite{petereit2012hybridastar}, and \cite{banfi2020planningundermalicious}. Hybrid A* produces smooth, drivable paths which obey the robot kinematics, and has previously been proven successful in practical robotic systems. The local goal point for hybrid A* is generated by our multiple-hypothesis planner using the approach described in section \ref{sec:local_goal}. We note that the multiple-hypothesis planner is agnostic to the specific choice of local planner.

We run the hybrid A* search using the current obstacle estimate means as obstacles. Since hybrid A* is a local planner, it operates only in a short range where sensor noise is minimal and estimate means are reliable. We check for obstacle collisions by bloating each obstacle's radius by half the robot width, and then modeling the robot as a point robot.

As mentioned in \cite{dolgov2010pathplanning}, a common pitfall with the hybrid A* planning algorithm is that the returned path can contain unnecessary side-to-side turns, due to the fixed motion primitives used by the algorithm, and can also pass very close to obstacles due to the search finding the shortest possible path. Therefore, as in Dolgov \etal{} \cite{dolgov2010pathplanning}, we use a gradient descent path smoother, optimizing for the distance of the path from nearby obstacles as well as the smoothness of the path.

\subsubsection{Baseline global planner}

In our experiments, we compare our multiple-hypothesis planner to a baseline global planner which uses a 2D A* search to find the shortest path to the goal, but does not account for the uncertainty in obstacle estimates. The baseline A* planner performs a search in 2D over a fixed-resolution grid. The search does not pass through grid cells which are in collision with an estimated obstacle mean. This search ignores the robot kinematics, and therefore is much quicker to execute than the 3D hybrid A* search. This lower computation time is required since the global planner searches over a much longer distance to the goal, compared to the local planner which plans a path only a few meters in length. After the 2D A* global planner computes a path to the global goal, the point $\planaheaddist$ distance ahead of the robot on this path is selected and used as the goal for the hybrid A* local planner, in order to find a short-range kinematically feasible path for the robot. We use the same local planner for the baseline as we do for our multiple-hypothesis planner, in order to perform a controlled comparison.

\subsection{Quantitative analysis}

\begin{figure*}
    \centering
    \includegraphics[width=\textwidth]{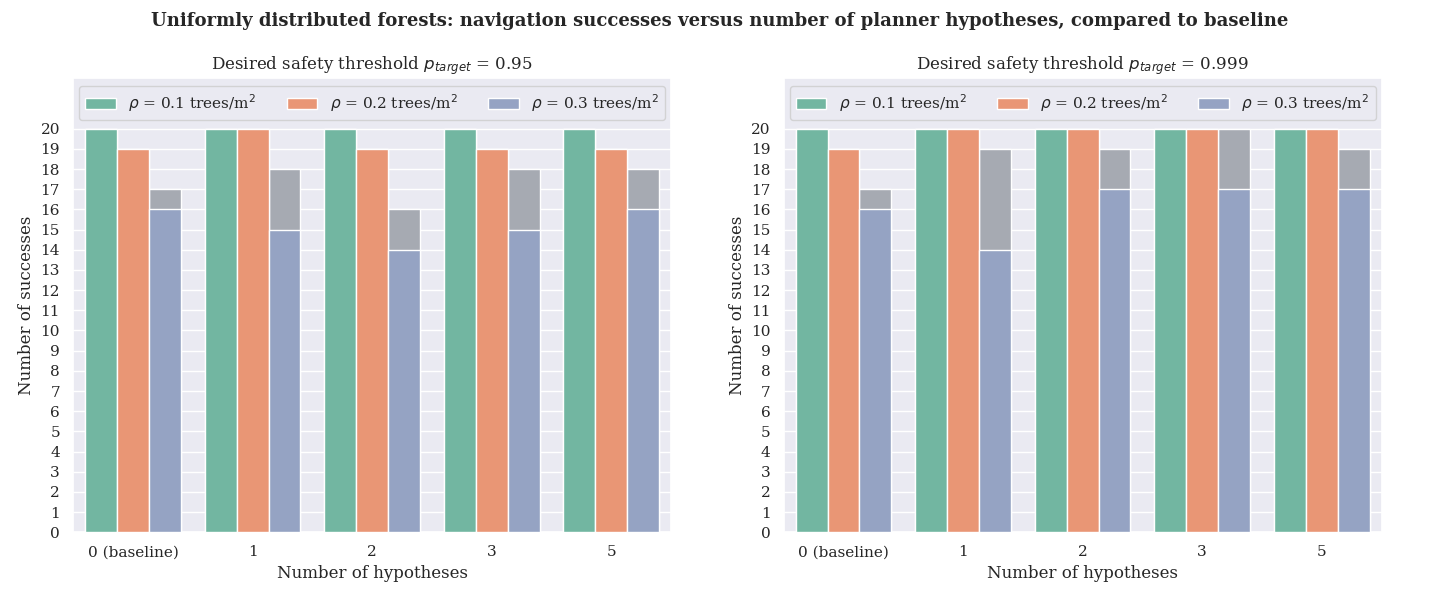}
    \caption{Results from simulating a differential drive robot navigating through 20 randomly generated Poisson forests with \emph{uniformly distributed tree positions}. The straight-line distance from the start to the goal, ignoring obstacles, is 40 meters. We run our experiments with three different settings for $\rho$, the density of trees in the forest (indicated by the three different colors), and with different maximum numbers of hypotheses $\nhyp$ allowed for the planner. Zero hypotheses indicates the 2D A* baseline global planner. The height of the bars indicate the number of generated forests in which the robot did not crash into an obstacle while navigating to the goal. The colored portions of the bars indicate the number of forests in which the robot successfully reached the goal. The gray portions of the bars indicate runs where the robot stopped early (without crashing into an obstacle), either due to the global planner believing there is no safe path to the goal available, or due to hybrid A* planner failure. \textbf{Left:} Results using a target safety probability of $\desiredprob=0.95$ for the multiple hypothesis planner. \textbf{Right:} Results using $\desiredprob=0.999$.} 
    \label{fig:exp_uniform}
\end{figure*}

\begin{figure*}
    \centering
    \includegraphics[width=\textwidth]{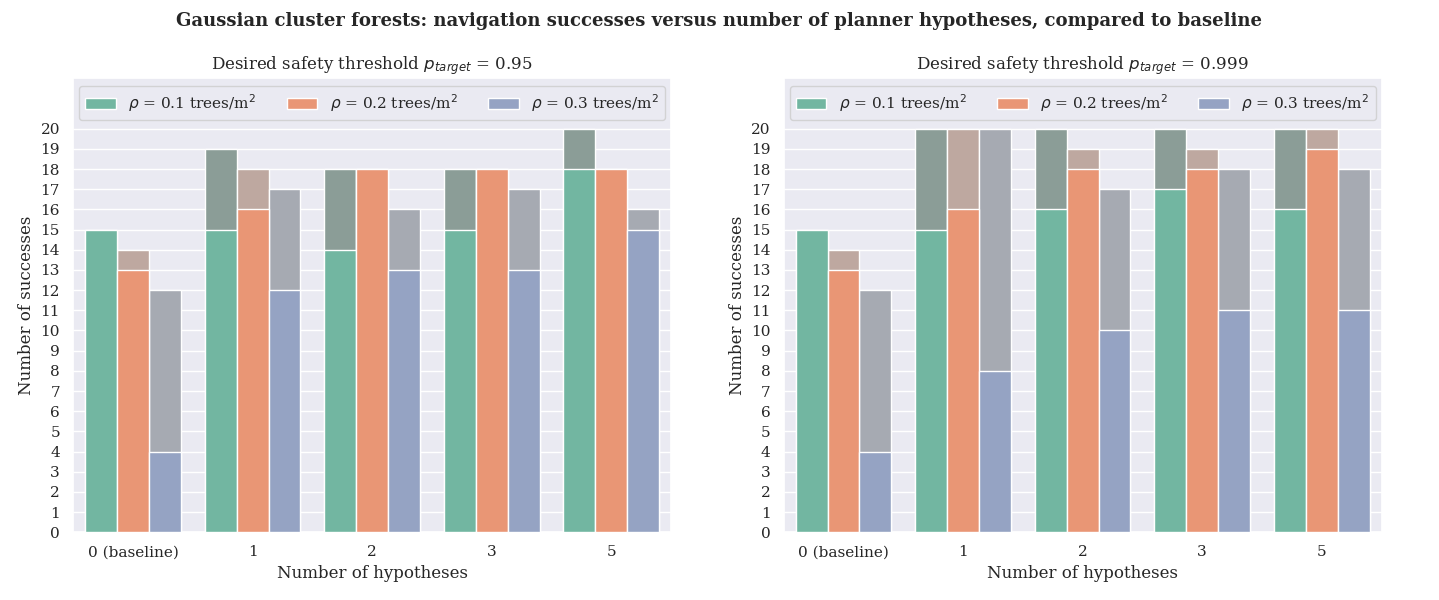}
    \caption{Results of planner successes in forests which contain Gaussian-distributed tree clusters between the robot's starting position and the goal point. Subplot arrangement is the same as in Figure \ref{fig:exp_uniform}. \textbf{Left:} Results using a target safety probability of $\desiredprob=0.95$ for the multiple hypothesis planner. \textbf{Right:} Results using $\desiredprob=0.999$} 
    \label{fig:exp_clusters}
\end{figure*}

We experimentally evaluate our multiple hypothesis planner in randomly generated forests, comparing it to the baseline 2D A* global planner. In order to study our planner's effectiveness in different types of environments, we vary the forest density and distribution of trees (including forests with uniformly distributed trees and forests containing nonuniform clusters of trees, as described in Section \ref{sec:sim_forests}). We generate 20 random forests for each combination of tree density and distribution, and run the multiple hypothesis planner varying the parameters $\nhyp$ and $\desiredprob$. The simulated robot re-plans iteratively at a rate of 1Hz, and obtains new detections at a rate of 2Hz. The robot is able to detect all non-occluded obstacles within a maximum range of 20 meters that lie in a front-facing sensor field of view of 110 degrees. We set the graph construction to ignore estimated obstacles at a distance of greater than 15 meters, as we find that near the sensor's maximum range, obstacle estimates have received very few associated detections (or only a single detection), and therefore contain too high uncertainty to be useful for planning. The robot's speed varies between $1~\si{m\per s}$ and $5~\si{m\per s}$, driving more slowly when an obstacle is nearby. Finally, the planner internally adds a barrier of artificial obstacles around the limits of the environment in order to keep the robot within the boundaries of the simulated environment.

Figure \ref{fig:exp_uniform} shows simulation results from the uniformly distributed forest environments, with multiple numbers of maximum hypotheses $\nhyp$ and two different settings for $\desiredprob$. We plot the number of cases in which the robot successfully reached the goal, indicated by the colored bars. The \emph{gray portions of the bars} indicate cases where the robot was unable to find a path to the goal and stopped, rather than crashing into an obstacle. This can occur either when the hybrid A* local planner fails to find a path to the local goal, or the global planner fails to find a path to the global goal. In the case of the multiple hypothesis global planner, this failure indicates that the planner believes no path above safety $\desiredprob$ exists in the \graphname{}.

The overall success rate of the planner is high in the uniform forest environments, even when using the baseline global planner, as the robot can generally avoid uniformly distributed obstacles by driving mostly in a straight line towards the goal, making slight deviations around an obstacle when needed. Still, we observe a higher rate of crashes using the baseline planner in higher density forests. In the highest density forests, with $\rho=0.3~\si{tree\per m^2}$, the baseline crashes 3 times out of 20 and stops once, compared to 1 crash for the medium density and no crashes in the lowest density forest. These findings show that denser obstacles, even in the relatively simple uniformly distributed forests, are still more challenging for navigation.

Our multiple hypothesis planner with $\desiredprob=0.95$ closely matches the performance of the baseline, likely due to the already low rate of crashes. In the highest density forests, the 5-hypothesis planner crashes twice and stops twice, out of 20 generated forests.  With $\desiredprob=0.999$, we see that the multiple hypothesis planner achieves a higher rate of safe navigation compared to the baseline. The planner always safely reaches the goal in the low and medium density forests, and in the high density case, the planner never crashes with $\nhyp=3$, and crashes once out of 20 forests for the 1, 2, and 5-hypotheses cases.

We then analyze the performance of our planner in non-uniformly distributed forests, containing clusters of obstacles that the robot should avoid, with results shown in Figure \ref{fig:exp_clusters}. We observe that in these environments, the 2D A* baseline global planner performs poorly, reaching the goal safely only 4 out of 20 times in the  forests, and overall demonstrating a much lower rate of safe navigation as compared to experiments in the uniform forests.

In these more challenging, non-uniform forests, the multiple-hypothesis planner significantly increases the rate at which the robot is able to safely reach the goal. At $\desiredprob=0.95$, using just a single hypothesis in the high density forests triples the number of navigation successes compared to the baseline, from 4 to 12. Using 5 hypotheses further increases the number of successes to 15. We also see improvements in the multiple hypothesis planner's success rate, compared to the baseline, in the low and medium density forests. 

Increasing the target safety probability to $\desiredprob=0.999$ further decreases the number of crashes in the non-uniformly distributed forests. At this setting, the 5-hypothesis planner never crashes in the low and medium density cluster forests, and crashes twice in the high density forests as compared to four times for the $\desiredprob=0.95$ 5-hypothesis planner. However, the increased $\desiredprob$ does cause the robot to stop more often, due to being unable to find a safe enough path to the goal. The planner stops most often in the high density forests, with the 5-hypothesis planner succeeding 11 times, stopping 7 times, and crashing 2 times in the high density forests. This effect is intuitive for the higher safety threshold, as the planner will be more conservative overall. We note that depending on the application, a robot stopping when unable to find a path is likely more desirable than crashing into an obstacle. Overall, at both $\desiredprob$ settings, the multiple hypothesis planner significantly outperforms the baseline at all forest densities, increasing the number of successful navigation runs and/or reducing the number of crashes by stopping before hitting obstacles.

We note that the actual observed success rate of the planner often does not match the target safety probability $\desiredprob$ exactly. The planner targets a safety probability $\desiredprob$ for each planned path, but since multiple re-planning iterations are required to reach the goal, the overall likelihood that \emph{all} of the planned paths are safe is below $\desiredprob$. Additionally, the other components of the planning pipeline, including measurement data association and the local hybrid A* planner, can occasionally introduce other failures that cause crashes (for example, hybrid A* planning too close to an obstacle due to approximations when checking collisions during the motion primitives). Still, our findings do show increased safety for the higher setting of $\desiredprob$, indicating that the path safeties approximated using our safety probability model and high-level graph representation do in fact correspond to higher rates of the robot actually reaching its goal. 

%% file: experiments_subsections/qualitative.tex

\subsection{Qualitative analysis}

\begin{figure*}
    \includegraphics[width=\textwidth]{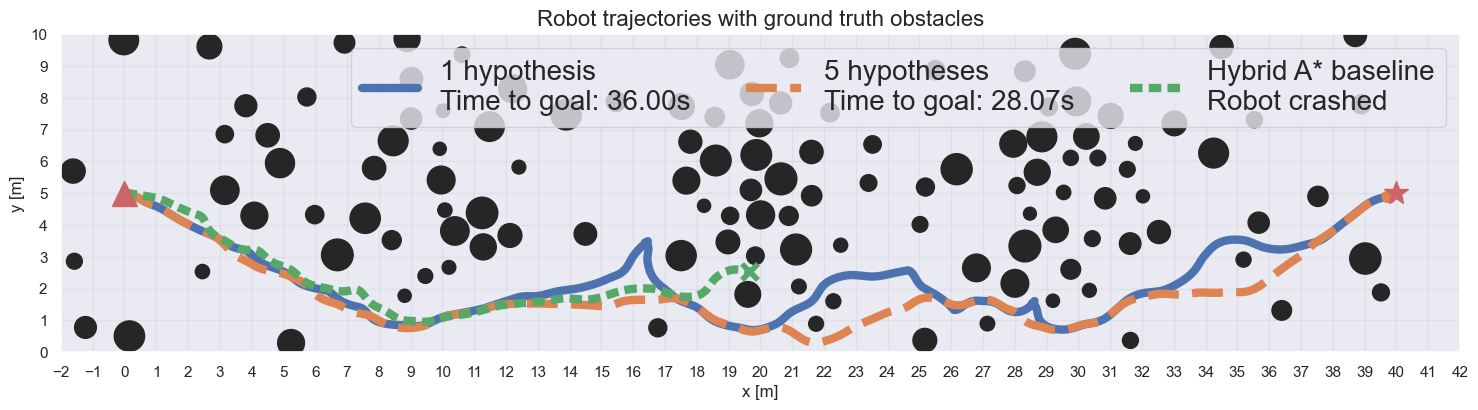}
    \caption{Behavior of the 1-hypothesis and 5-hypothesis planners, plotted with the hybrid A* planner trajectory, in a high-density randomly generated forest. The 1- and 5-hypothesis planners use $\desiredprob=0.95$.}
    \label{fig:qual_base_1hyp_5hyp}
\end{figure*}

We present several visualizations of the planner's behavior in order to demonstrate the effects of using multiple plan hypotheses, as well as of adjusting the planner's path safety threshold.

Figure \ref{fig:qual_base_1hyp_5hyp} shows the trajectories taken by the robot through a randomly generated forest (containing nonuniformly distributed obstacle clusters at the highest obstacle density setting), using the 1-hypothesis, 5-hypothesis, and hybrid A* baseline planners. The hybrid A* planner crashes into an obstacle, as it attempts to navigate between a pair of close-together obstacles. The single hypothesis planner successfully reaches the goal, but backtracks and changes course multiple times, as it reacts to newly obtained sensor measurements of nearby obstacles. The 5-hypothesis planner reaches the goal with the smoothest path overall, as it accounts for further-away, uncertain obstacles when planning.

\begin{figure}
    \subfloat[Zoomed-in view comparing paths generated with the planner using 1 hypothesis versus 5 hypotheses.\label{fig:qual_1hyp_vs_5hyp_top}]{\includegraphics[width=\columnwidth]{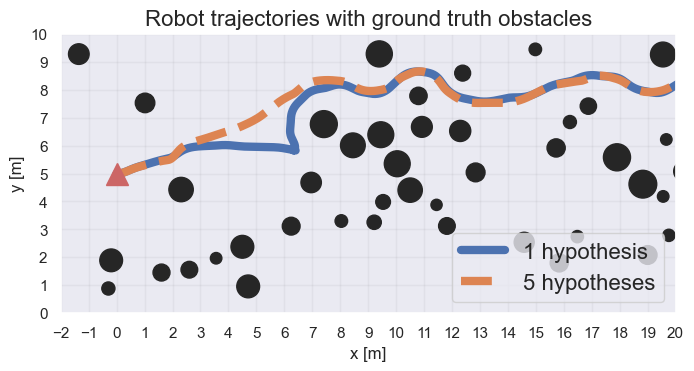}}

    \subfloat[Navigation graph constructed by the planner, and the path computed by the single-hypothesis planner.\label{fig:qual_1hyp_vs_5hyp_mid}]{\includegraphics[width=\columnwidth]{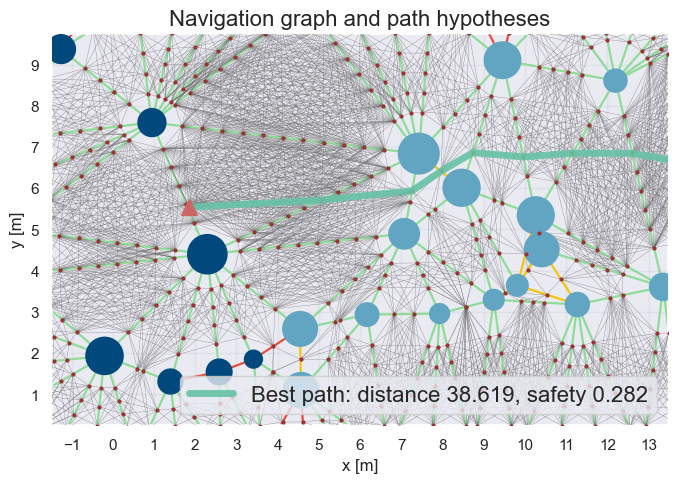}}

    \subfloat[Multiple path hypotheses generated by the multiple-hypothesis planner with up to 5 maximum hypotheses allowed, showing the best path (green) and two alternative hypotheses that were considered by the planner.\label{fig:qual_1hyp_vs_5hyp_bottom}]{\includegraphics[width=\columnwidth]{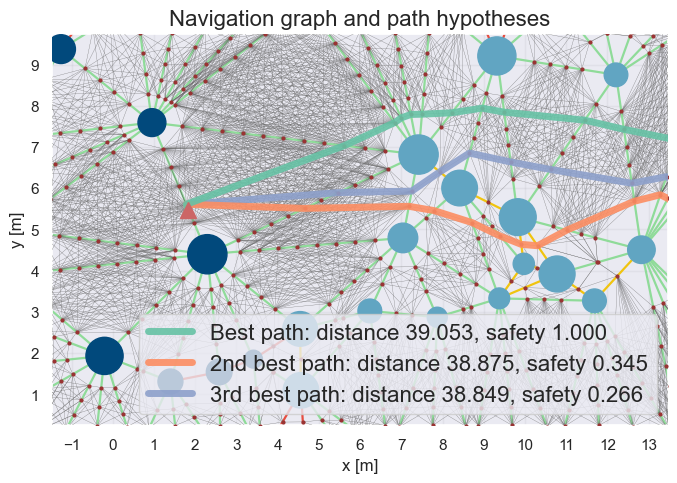}}
    
    \caption{Comparison between the behaviors of the 1-hypothesis and 5-hypothesis versions of our planner. Both planners use $\desiredprob=0.95$.}
    \label{fig:qual_1hyp_vs_5hyp}
\end{figure}

Figure \ref{fig:qual_1hyp_vs_5hyp} zooms in on an example case to demonstrate how the planner behaves differently when using 1 hypothesis versus 5 hypotheses. The robot trajectories, plotted in Figure \ref{fig:qual_1hyp_vs_5hyp_top} show that the 1-hypothesis planner finds a safe path around the cluster of obstacles ahead, but has to closely approach these obstacles before determining there is no path through them with a 95\% likelihood of being safe. In contrast, the 5-hypothesis planner identifies ahead of time that a safer, but slightly longer distance, path exists by traveling around the cluster. Figures \ref{fig:qual_1hyp_vs_5hyp_mid} and \ref{fig:qual_1hyp_vs_5hyp_bottom} illustrate the navigation graph constructed by the planner, and the path hypotheses considered by the 1- and 5-hypothesis versions of the planner. The 1-hypothesis planner computes only the shortest path to the goal in the navigation graph, which passes through an uncertain, narrow gap between two obstacles, ultimately leading to the robot needing to backtrack away from these obstacles. In contrast, the 5-hypothesis planner identifies a path that takes a slightly longer route around these obstacles, but has a safety probability near 1.0. The multiple-hypothesis planner chooses this as the best path, trading off distance for a much safer path.

\begin{figure}
    \subfloat[A case where the 5-hypothesis planner crashes with the lower setting for the safety probability threshold $\desiredprob$, but finds a safe route to the goal using the higher setting.\label{fig:qual_5hyp_low_high_safety_top}]{\includegraphics[width=\columnwidth]{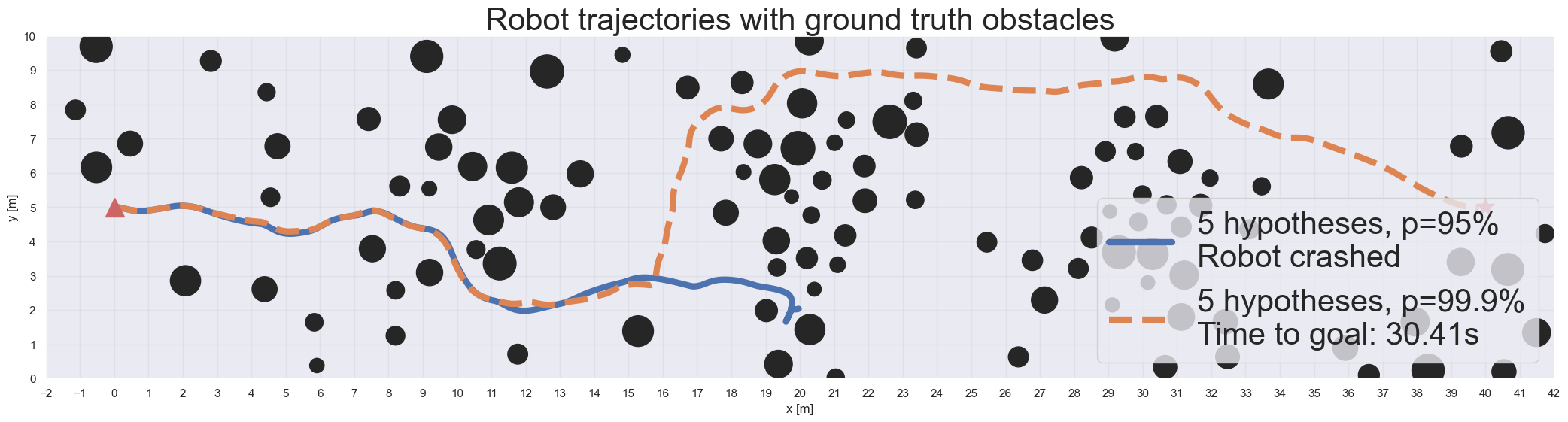}}
    
    \subfloat[A case where the planner reaches the goal successfully with both settings of $\desiredprob$, but takes a longer detour when using the higher setting.\label{fig:qual_5hyp_low_high_safety_mid}]{\includegraphics[width=\columnwidth]{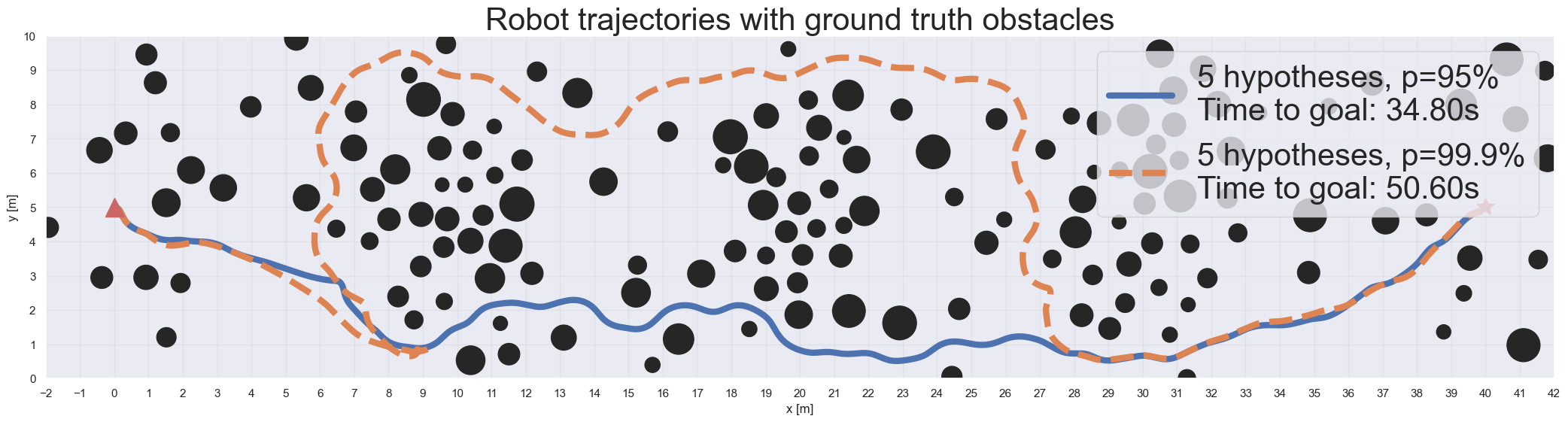}}
    
    \subfloat[A case where the planner reaches the goal with the lower setting for $\desiredprob$, but stops when using the higher setting, due to the planner being unable to find a path that satisfies the safety threshold.\label{fig:qual_5hyp_low_high_safety_bottom}]{\includegraphics[width=\columnwidth]{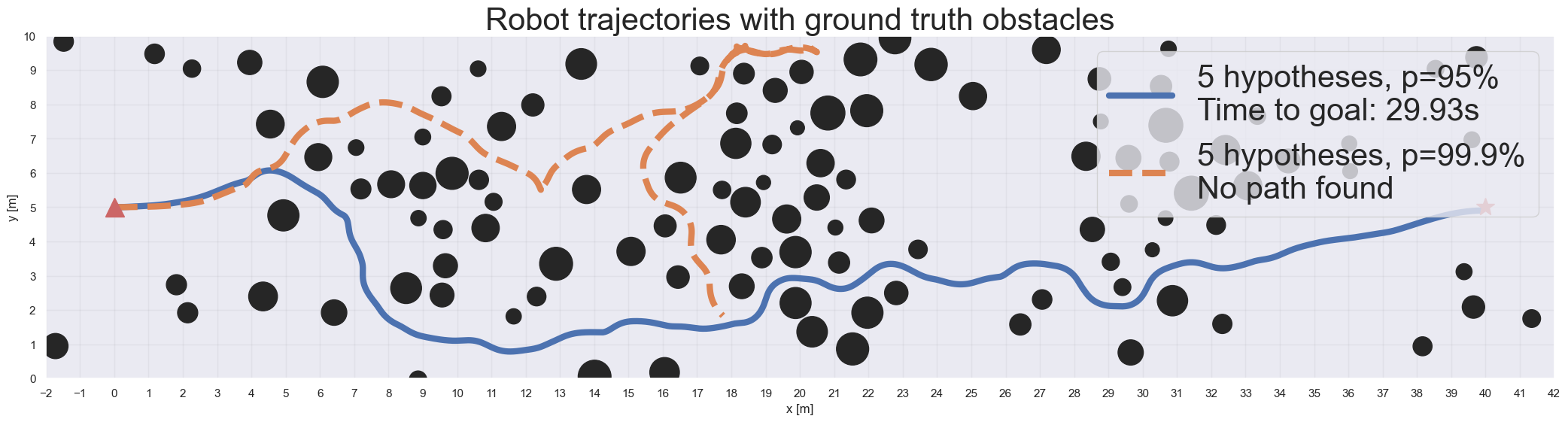}}
    \caption{Multiple cases showing how the behavior of the 5-hypothesis planner changes when the planner searches for a path with safety probability $\desiredprob=0.95$ versus $\desiredprob=0.999$.}
    \label{fig:qual_5hyp_low_high_safety}
\end{figure}

Finally, Figure \ref{fig:qual_5hyp_low_high_safety} demonstrates the effects of changing the desired path safety probability $\desiredprob$, showing the behavior of the 5-hypothesis planner with a desired path safety of $\desiredprob=0.95$ versus with $\desiredprob=0.999$, in three different forest environments. In Figure \ref{fig:qual_5hyp_low_high_safety_top}, the lower-safety threshold planner crashes as it attempts to navigate a gap between several obstacles, while the higher-safety planner identifies this unsafe region and avoids it. However in other cases, the more conservative behavior of the higher safety threshold planner can cause the robot to take longer to reach the goal, as seen in Figure \ref{fig:qual_5hyp_low_high_safety_mid}, or to fail to find a path when a safe path does in fact exist, as seen in Figure \ref{fig:qual_5hyp_low_high_safety_bottom}. These multiple cases show that the safety threshold $\desiredprob$ significantly affects the planned path, and should be set depending on the desired trade-off of safety with the time to reach the goal.

%% file: conclusion.tex

\section{Conclusion}

In this paper, we presented a method for path planning through complex, obstacle-dense environments using noisy object detections, motivated by the use of machine learning-based object detectors to enable robots to plan around distant obstacles. Our approach constructs a probabilistic graph representation of obstacles; we then plan multiple candidate paths through the graph, accounting for path safety and expected distance to optimize the best route to the goal. Simulation experiments in generated forest environments demonstrated that our multiple hypothesis planner enables a differential drive robot to reach its goal much more safely in complex, obstacle-dense forests as compared to a baseline global planner. 

Our graph representation provides a probabilistic framework that could be extended with other sources of uncertain information. In particular, the evaluation metric for the multiple path hypotheses is a promising area for future improvement, building upon the current path cost function based on safety and distance-to-goal. Including information gain as a weight in the path evaluation, as in \cite{yutaoDeepSemanticHPPC}, would encourage the robot to follow paths that allow it to view previously unobserved regions of the environment, or to approach uncertain obstacles in order to improve its estimate certainty by collecting additional measurements. Image semantics are another promising source of information \cite{yutaoDeepSemanticHPPC, ryll2020semantic}, and could be used to weight candidate paths by recognizing obstacle-dense regions from long range using semantic masks.